%% file: neurips_2026.tex
\documentclass{article}

\PassOptionsToPackage{numbers}{natbib}

\usepackage[main, final]{neurips_2026}

\usepackage[utf8]{inputenc} %
\usepackage[T1]{fontenc}    %
\usepackage{hyperref}       %
\usepackage{url}            %
\usepackage{booktabs}       %
\usepackage{amsfonts}       %
\usepackage{nicefrac}       %
\usepackage{microtype}      %
\usepackage{xcolor}         %

\usepackage{soul}
\usepackage{todonotes}
\usepackage{amsmath}
\usepackage{float}
\usepackage{placeins}
\usepackage{subcaption}
\usepackage{enumitem}

\newif\ifshowedits \showeditsfalse

\usepackage{pdfcomment}

\title{Event-Grounded Sparse Autoencoders for Vision-Language-Action Policies}

\author{%
  Xinchen~Jin \quad Aditya~Chatterjee \quad Pranav~Kumar \quad Rohan~Paleja \\
  Department of Computer Science, Purdue University \\
  West Lafayette, IN 47907 \\
  \texttt{\{jin548, chatte59, kumar649, rpaleja\}@purdue.edu}
}
\begin{document}

\maketitle

\begin{abstract}
Vision-Language-Action (VLA) policies translate language and visual inputs into robot actions, where their hidden representations directly shape closed-loop behavior. However, mechanistic interpretability tools from language and vision-language models do not transfer cleanly to VLAs: outputs are robot actions rather than human-readable tokens, and validating internal changes requires expensive closed-loop rollouts. We propose an event-grounded interpretability pipeline that anchors SAE feature analysis to behavioral events rather than text contexts.
We extract keyframes from end-effector trajectories and cluster them within each task using visual, robot-state, and temporal cues. The resulting events guide SAE feature ranking, with optional VLM annotations providing descriptions of the events. To our knowledge, our
pipeline is among the first to ground SAE-based VLA analysis in closed-loop behavioral events. Across two simulation architectures and a real-robot study, event-grounded ranking yields the strongest causal effects on OpenVLA and transfers to the continuous action chunks of $\pi_{0.5}$. SAE is a sparse but imperfect intervention basis: usability varies with architecture and intervention site, and aggressive intervention reveals safety and interpretability limits.
Overall, event-grounded SAE analysis emerges as a practical starting point for behavior-anchored VLA interpretability, motivating future work on SAE analysis beyond end-effector motion, finer-grained closed-loop evaluation, and safe interventions for high-stakes VLA deployments.
\end{abstract}

\section{Introduction}
\label{sec:intro}

Vision-Language-Action (VLA) models map language instructions and visual
observations to robot actions and are increasingly deployed in closed-loop
physical settings~\citep{kim2024openvla, black2024pi0, intelligence2025pi05}.
Because VLA outputs are actions that directly drive robot motion rather
than the human-readable tokens emitted by language and vision-language
models, hidden representations in a VLA are not only useful for predicting next
actions but also directly shape what the robot does next: small internal
changes can translate into different closed-loop trajectories, contact
events, or failure modes.
What these representations encode therefore matters both for interpretability research and for safely deploying VLA policies in the real world~\citep{icct-rss-22, paleja2021utility}.

Mechanistic interpretability has matured in language and vision-language
models, where intermediate representations can be related to human-readable
tokens through tools such as the logit lens or sparse autoencoder (SAE)
feature decoding~\citep{nostalgebraist2020logitlens, gao2025scaling}.
These methods exploit two properties of language model outputs:
(i) candidate features can be named by projecting internal directions into
vocabulary space~\citep{geva2022transformer, dar2023analyzing} or by
inspecting their high-activating text contexts~\citep{bricken2023towards,
cunningham2023sparse}; (ii) text-based activation examples enable scalable
automated explanation and validation~\citep{bills2023language,
paulo2025automatically}.
Neither property transfers cleanly to VLAs. First, feature naming
is harder: the logit lens does not transfer, since VLA actions live in
a control space rather than the language vocabulary. Second, behavioral
validation is harder: closed-loop evaluation is expensive, and an
action change on its own does not distinguish noise from meaningful
effect. Unlike a language model, where the output token gives a direct semantic
readout, a VLA produces a semantic outcome only after many correct
downstream steps. These constraints motivate grounding VLA feature analysis in recurring behavioral events from rollouts, rather than in projections to the language vocabulary.
Recent work has begun to study the internal representations of VLA
policies~\citep{swann2026sparse, haon2025mechanistic, li2025task, khan2025controlling, grant2026not, buurmeijer2026observing}.
In the concurrent work of \citet{swann2026sparse}, candidate SAE features
are first selected from activation statistics and then labeled by
manually aligning their activation traces with rollout video. Behavior
thus enters at the interpretation stage, where manual labeling
requires per-feature effort and visual co-occurrence
does not by itself establish a feature's causal
role. Our pipeline instead scores every alive SAE feature against external
behavioral events, rather than a subset selected from activation statistics.

\begin{figure}[!htbp]
    \centering
    \includegraphics[width=\linewidth]{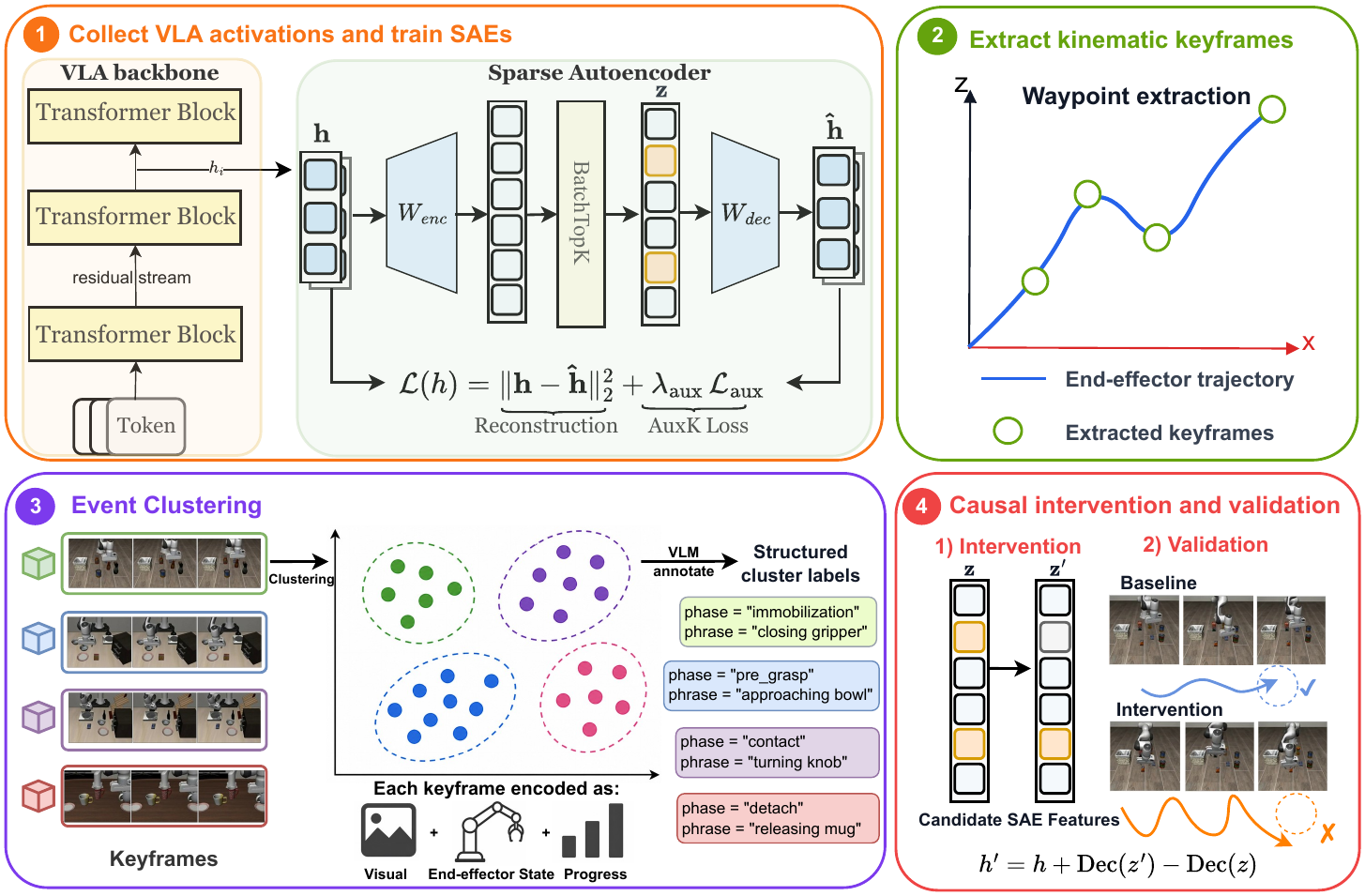}
    \caption{This figure shows the 4 stages of the event-grounded SAE
    pipeline: (1) SAE training (Sec.~\ref{sec:sae-training}); (2)
    kinematic keyframe extraction (Sec.~\ref{sec:keyframe-extraction});
    (3) task-local event clustering with VLM annotation
    (Sec.~\ref{sec:clustering}); (4) closed-loop validation
    through residual-preserving latent edits
    (Sec.~\ref{sec:intervention}).}
    \label{fig:pipeline}
\end{figure}

We use SAEs because they decompose superposed activations~\citep{elhage2022toy} into a sparse basis whose features can be enumerated and scored individually~\citep{bricken2023towards, templeton2024scaling}, and because SAE training is unsupervised. To select features worth testing, our pipeline uses SAE-independent behavioral signals (Figure~\ref{fig:pipeline}): we extract kinematic keyframes via Automatic Waypoint Extraction~\citep{shi2023waypointbased}, cluster into events, score features against clusters, and validate features through closed-loop interventions. Extraction and scoring are automatic, so the pipeline scales beyond per-feature manual labeling.
\noindent Our contributions are:
\begin{itemize}[leftmargin=*, topsep=0pt, itemsep=2pt, parsep=0pt]
    \item \textbf{An automated, event-grounded SAE pipeline for VLAs.}
    We extract kinematic keyframes from rollouts, cluster them into
    task-local events, and score every alive SAE feature against these
    events without per-feature manual labeling. The event anchors are
    independent of SAE activations; feature ranking then scores SAE
    activations conditioned on these external behavioral events.

   \item \textbf{Architecture- and site-dependent intervention behavior.}
    The same pipeline yields qualitatively different results: on OpenVLA, individual event-aligned features measurably drop closed-loop success; on $\pi_{0.5}$, the PaliGemma VLA backbone barely responds to single-feature edits while the action expert collapses under nearly any score-based ranking. Intervention sites are architecture-specific and do not transfer directly.

    \item \textbf{Behavioral evaluation beyond binary success.}
    Across simulation and a real-robot study, 
    we show that single-feature interventions can change robot motion
    in ways that binary success rates do not capture. Although these
    interventions cannot cleanly isolate causal variables, the findings motivate
    finer-grained evaluation metrics and methods.
\end{itemize}

\section{Background and Related Work}

\textbf{Mechanistic Interpretability of LLMs and VLMs.}
In LLMs, prior work has studied how information is stored and
transformed across layers and components~\citep{meng2022locating,
geva2023dissecting}, and used tools such as the logit lens to inspect
how predictions evolve over depth by projecting intermediate
activations through the unembedding
matrix~\citep{nostalgebraist2020logitlens, geva2022transformer,
dar2023analyzing}. These methods rely on the discrete, human-readable
output vocabulary, which makes intermediate representations relatively
easy to inspect and validate. The same advantage largely carries over
to text-generating VLMs, and similar analyses have
begun to uncover multimodal circuits and
representations~\citep{neo2025towards, phukan2025beyond}.

\textbf{Sparse Autoencoders.}
A line of interpretability work from Anthropic identified superposition
in neural networks~\citep{elhage2022toy}, introduced sparse
autoencoders (SAEs) as a means of recovering more interpretable latent
features from activations~\citep{bricken2023towards}, and later scaled
this framework to large language models, showing that SAE features can
capture abstract, multilingual, and safety-relevant concepts and
support causal interventions~\citep{templeton2024scaling}. A growing body of work has applied SAEs to interpret LLMs more broadly~\citep{cunningham2023sparse, bills2023language, chanin2025a,
paulo2025automatically}, and similar ideas have recently been extended
to VLMs~\citep{lou2025saev, pach2025sparse,
papadimitriou2025interpreting}. In both settings, feature semantics
can often be inferred from activating text or images and from model outputs, and feature validation can
draw on large, semantically annotated corpora.

\textbf{Mechanistic Interpretability of VLAs.}
Applying these tools to VLAs raises new challenges, since outputs are
actions and intervention quality must be assessed through closed-loop
behavior. Moreover, studies of robot policies suggest that successful task
execution need not reflect generalizable task
understanding~\citep{he2026demystifying, fang2026vision}. Policy representations therefore need not form clean, transferable semantic abstractions,
so perfectly disentangled SAE features should not be assumed. Non-SAE approaches include
FFN-direction analysis for physical quantities such as end-effector
speed and height~\citep{haon2025mechanistic} and text-latent
intervention for task redirection~\citep{li2025task}. Several recent
works apply SAEs to VLA policies: \citet{khan2025controlling} derive SAE steering directions from input pairs that differ in the target action, and
\citet{grant2026not} find that per-token SAE processing is essential
for preserving VLA action fidelity. \citet{buurmeijer2026observing}
take a complementary linear-probe approach to observe and steer VLA
features. Most directly comparable to our pipeline, concurrent work
by~\citet{swann2026sparse}
classifies features as general or memorized using a classifier
trained on activation-derived statistics. Unlike these works, which start from a prespecified target behavior or from SAE activation statistics, our
pipeline scores SAE features against SAE-independent behavioral
events extracted from rollouts.

\section{Method}
\label{sec:method}
Our pipeline turns closed-loop VLA rollouts into an SAE-based,
event-grounded basis for causal feature analysis
(Figure~\ref{fig:pipeline}). Each upstream stage is anchored by a
signal independent of the SAE activations being analyzed: candidate
features are proposed from kinematic key moments that align with human
intuition about a task's salient steps. The remainder of this section
covers SAE training (Section~\ref{sec:sae-training}), kinematic keyframe
extraction (Section~\ref{sec:keyframe-extraction}), event clustering
(Section~\ref{sec:clustering}), feature ranking
(Section~\ref{sec:feature-ranking}), and closed-loop intervention
(Section~\ref{sec:intervention}), together giving a way to discover,
score, and causally test SAE features without manual per-feature
labeling.
\subsection{SAE Training and Fidelity Evaluation}
\label{sec:sae-training}

SAEs decompose dense hidden activations into a higher-dimensional
sparse basis whose features are intended to be more interpretable than
individual neurons~\citep{bricken2023towards, elhage2022toy} (Stage (1)
of Figure~\ref{fig:pipeline}). Given a hidden state
$\mathbf{h} \in \mathbb{R}^{d}$, the encoder produces a sparse code
$\mathbf{z} = \mathrm{Enc}(\mathbf{h}) = \mathrm{BatchTopK}\!\left(W_{\mathrm{enc}}\mathbf{h}
+ \mathbf{b}_{\mathrm{enc}}\right) \in \mathbb{R}^{m}$ with $m \geq d$.
The $\mathrm{BatchTopK}$ operator retains only the top $kB$
pre-activations and zeros the rest, where $B$ is the minibatch size
and $k$ is the average per-example active budget~\citep{bussmann2024batchtopk}.
This architectural top-$k$ constraint replaces the $L_1$ sparsity
penalty of earlier SAEs. The decoder reconstructs as $\hat{\mathbf{h}}
= \mathrm{Dec}(\mathbf{z}) = W_{\mathrm{dec}} \mathbf{z} + \mathbf{b}_{\mathrm{dec}}$, and we
train both jointly to minimize
$\mathcal{L}_{\mathrm{SAE}} = \tfrac{1}{B} \sum_{i=1}^{B}
\|\mathbf{h}^{(i)} - \hat{\mathbf{h}}^{(i)}\|_2^2 + \lambda_{\mathrm{aux}}
\mathcal{L}_{\mathrm{aux}}$,
where $\mathcal{L}_{\mathrm{aux}}$ is the AuxK loss that revives dead
features~\citep{gao2025scaling}.

We train one BatchTopK SAE per (architecture stream, suite, layer)
combination on per-token residual-stream activations from closed-loop
rollouts. Activations are not mean-pooled across tokens before training.
SAE quality is assessed through two criteria: offline diagnostics and
closed-loop behavioral fidelity. The offline diagnostics are the fraction of
variance explained (FVE), the alive fraction, and the average $L_0$. Behavioral
fidelity is the closed-loop success rate (SR) under a reconstruction-only hook
that replaces the hidden state at layer $\ell$ with its reconstruction
$\hat{\mathbf{h}}$. We call SR measured under
this hook \emph{Hooked SR}. Hooked SR selects the layer for
downstream discovery and intervention.

\subsection{Kinematic Keyframe Extraction}
\label{sec:keyframe-extraction}

We use Automatic Waypoint Extraction
(AWE)~\citep{shi2023waypointbased} to identify candidate event
anchors (Stage (2) of Figure~\ref{fig:pipeline}). AWE compresses a
trajectory into a sparse ordered subsequence of waypoints under a
reconstruction-error budget $\eta$, controlling the tradeoff between
waypoint sparsity and trajectory fidelity. Concretely, AWE picks the
smallest waypoint set such that linear interpolation between
consecutive waypoints stays within $\eta$ of the original trajectory
at every timestep. Given a rollout, we apply
AWE to the end-effector position trajectory and treat the resulting
waypoints as kinematic keyframes, candidate temporal anchors for
downstream event analysis. Because AWE depends only on end-effector
motion, the resulting keyframes are independent of SAE activations,
semantic labels, and feature scores, providing a
representation-independent proposal mechanism for candidate events.
For
each keyframe at timestep $t_i$, we construct a local observation
bundle: a five-frame image strip around $t_i$, a short video clip, and
metadata including task description, episode index, success label,
waypoint rank, and exact timestep. Figure~\ref{fig:event-cluster-examples}
shows four representative bundles.
One limitation is that using only end-effector motion limits the method
to kinematically salient events. A physical or semantic change that leaves
the end-effector trajectory unchanged, such as object slip or a switch
in which object is manipulated, would need proposal signals beyond
end-effector motion. Object motion, contact, and force proposals remain
future work.

\subsection{Event Clustering}
\label{sec:clustering}

To score SAE features against recurring event types rather than
individual keyframes, we group keyframe bundles into task-local event
clusters (Stage (3) of Figure~\ref{fig:pipeline}). Clustering on
end-effector position alone is insufficient: semantically distinct
events can occur at spatially nearby positions, such as a grasp and a
subsequent release when source and target locations are close, or two
passes through the same region in opposite directions. We therefore
form each cluster descriptor by concatenating a normalized visual
embedding of the image strip (v), a normalized robot state vector with
end-effector position and gripper status (s), and a normalized
temporal-progress scalar (p), with tunable weights controlling the
relative contribution of each component
(Appendix~\ref{app:clustering-hyperparams}).

We cluster descriptors task-locally with agglomerative clustering at a
fixed cosine-distance threshold, discarding clusters whose member
episodes cover less than half of the task's rollout episodes. The
retained clusters serve as event units for the feature ranking of
Section~\ref{sec:feature-ranking}.
For illustration, Figure~\ref{fig:cluster-space} visualizes the
resulting clusters in end-effector position space for four
$\pi_{0.5}$ LIBERO tasks.

To help interpret which behavioral events (e.g., approach,
contact, release) SAE features respond to, we additionally annotate
each cluster for OpenVLA on LIBERO with a short phrase and coarse phase
label. We obtain these annotations by querying a vision-language
model (VLM) on exemplar keyframe bundles, following the phase taxonomy
of \citet{chen2026robovlm}. Appendix~\ref{app:vlm-prompt} provides
the prompt template, and Appendix~\ref{app:event-feature-matrix}
shows the annotated clusters and event-aligned feature scores.
These labels support post-hoc inspection, but their text is not
used to compute feature-ranking scores.

\subsection{Feature Ranking Strategies}
\label{sec:feature-ranking}

A trained SAE provides a sparse feature basis, but interpreting which
of its many features are behaviorally meaningful is a separate
problem~\citep{shu-etal-2025-survey}. Because closed-loop intervention tests are
costly, we rank SAE features to select a small candidate set for evaluation.
The event clusters from Section~\ref{sec:clustering} provide a behavioral
reference independent of SAE activations.

\begin{figure}[!ht]
\centering
\begin{minipage}[t]{0.42\linewidth}
\centering
\includegraphics[width=\linewidth]{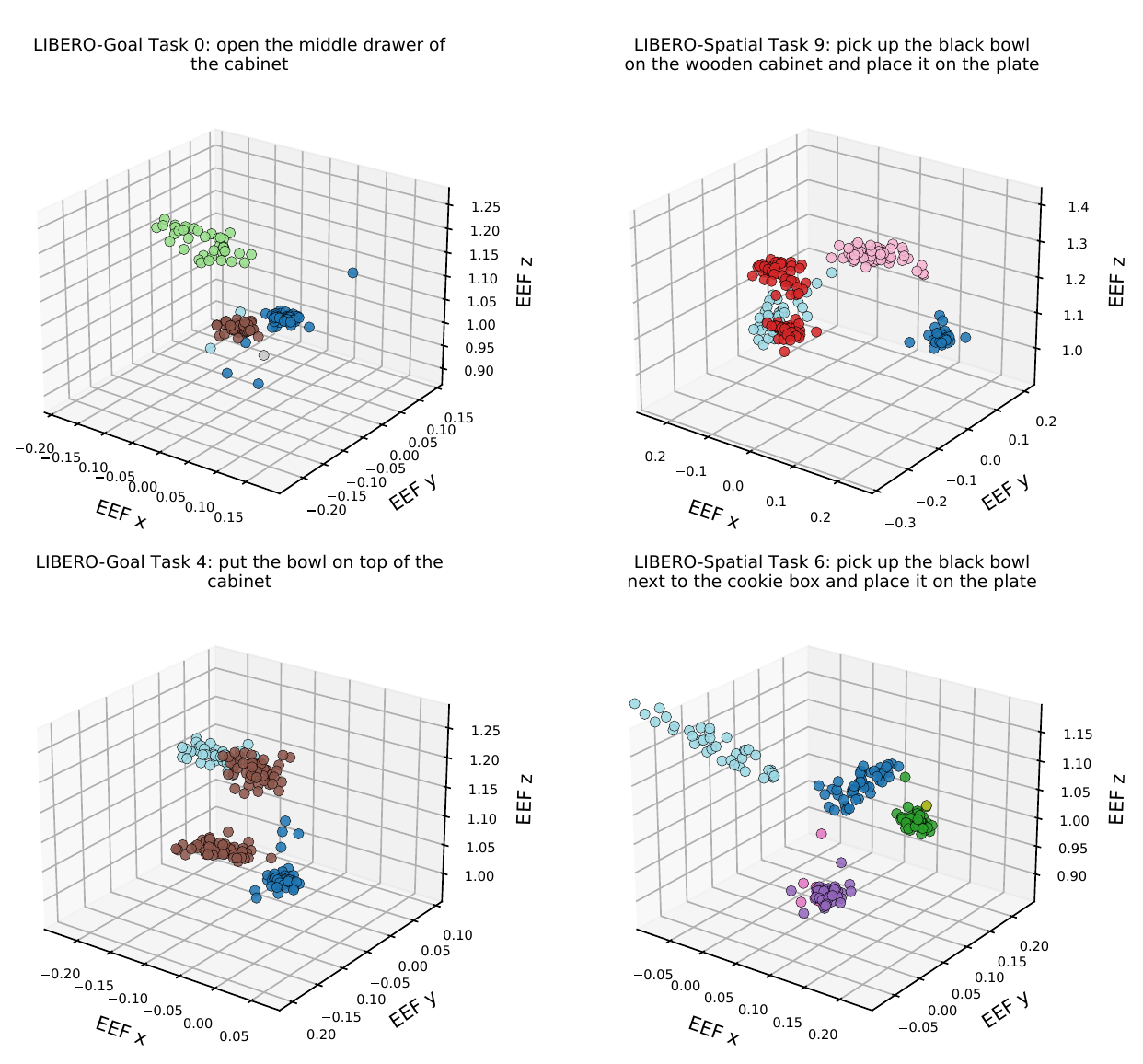}
\captionof{figure}{Example $\pi_{0.5}$ keyframe clusters in
end-effector position space for four LIBERO tasks. Each color is a
task-local event cluster.}
\label{fig:cluster-space}
\end{minipage}\hfill
\begin{minipage}[t]{0.55\linewidth}
\centering
\includegraphics[width=\linewidth]{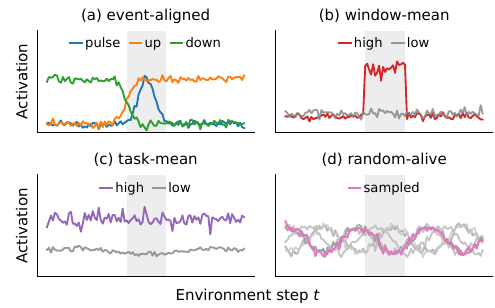}
\captionof{figure}{Schematic of the four ranking strategies. Each
panel shows the activation shape scored highly by the corresponding
ranking, with the gray band marking the event window centered at
$t_i$.}
\label{fig:ranking-schematic}
\end{minipage}
\end{figure}

We
compare three hypotheses about which features influence behavior: features whose
activation changes at recurring events (event-aligned ranking), features that
are highly active near these events (window-mean ranking), and features that
are highly active throughout the task (task-mean ranking). We also include a
random-alive ranking as a baseline. Figure~\ref{fig:ranking-schematic} illustrates these
rankings, and Appendix~\ref{app:ranking-formulas} gives the score formulas.

\textbf{Event-aligned ranking.}
For each keyframe in an event cluster at timestep $t_i$, the event-aligned score measures
how well a feature's activation trace inside a $\pm w$ window matches
one of three templates: a \emph{pulse} spiking at $t_i$ (e.g., a
momentary recognition), or a \emph{step-up}/\emph{step-down} marking
a state transition (e.g., gripper opening or closing). The window
mean is subtracted before matching, so the score reflects temporal
shape rather than baseline activation level. Window scores are averaged within
each event and rollout under the best-matching template, then averaged
across rollouts so that a single rollout contributing many nearby
keyframes does not dominate.

\textbf{Window-mean ranking.}
The window-mean ranking shares event windows with the event-aligned
ranking and scores features by their average activation within each
window. It tests whether features with high average activation near
events tend to be causally important.

\textbf{Task-mean ranking.}
The task-mean ranking scores features by their average activation
across all rollout timesteps in the suite. It tests whether features
with high average activation during a task (regardless of when, or near
which event) tend to be causally important.

\textbf{Random-alive ranking.}
As a sanity check, we also include a random ranking: we uniformly
sample $K$ features from the set of alive SAE features (those that
fire in at least one rollout of the suite), excluding any feature
already selected by the event-aligned, window-mean, or task-mean
rankings.
This exclusion ensures the random baseline measures the effect of
generic alive features, not features that happen to overlap with
score-based rankings. The ranking serves as a baseline: if the
score-based rankings did not outperform this baseline, they would carry
no useful signal.

\subsection{Intervention Modes}
\label{sec:intervention}

To test the causal role of selected SAE features, we intervene on
hidden representations during inference and measure the effect on
closed-loop behavior (Stage (4) of Figure~\ref{fig:pipeline}). We use
two intervention modes. The first is the reconstruction-only hook
from Section~\ref{sec:sae-training}, used to assess SAE fidelity.
The second, described below, is a residual-preserving latent edit
that modifies a chosen subset of features without disturbing the rest
of the hidden state. The goal is to steer behavior rather than
overwrite it completely.

\textbf{Residual-preserving latent edits.}
Let $\mathbf{h} \in \mathbb{R}^d$ be the hidden state at the intervention site
and $\mathbf{z} = \mathrm{Enc}(\mathbf{h}) \in \mathbb{R}^m$ its SAE code. We modify a
target subset of latent features $S$ by scaling them by a factor
$\alpha$:
\begin{equation}
z'_i =
\begin{cases}
\alpha\, z_i & i \in S, \\
z_i & i \notin S.
\end{cases}
\label{eq:latent-edit-z}
\end{equation}
With this parameterization, $\alpha = 0$ zeroes the targeted features,
$\alpha \in (0,1)$ softly suppresses them, $\alpha = 1$ recovers $\mathbf{z}$
(no edit), and $\alpha > 1$ amplifies them.
We update the hidden state by adding the change in the SAE reconstruction,
$\mathbf{h}' = \mathbf{h} + \mathrm{Dec}(\mathbf{z}') - \mathrm{Dec}(\mathbf{z})$.
For the linear decoder, the added perturbation is
$(\alpha - 1)\sum_{i \in S} z_i\mathbf{d}_i$, where $\mathbf{d}_i$ is the decoder
vector of feature $i$. Each selected feature therefore contributes in proportion
to its current activation. Adding only the reconstruction difference preserves the part of the original
hidden state not captured by the SAE.

\section{Experiments}
\label{sec:experiment}
\subsection{Setup and SAE Fidelity}
\label{sec:exp:setup}
\label{sec:exp-sae-reconstruction}

We evaluate the pipeline on OpenVLA~\citep{kim2024openvla} (discrete
action tokens) and $\pi_{0.5}$~\citep{intelligence2025pi05} (continuous
action chunks) on the four LIBERO
suites~\citep{liu2023libero}, with 50 closed-loop rollouts per task for activation collection.
We train one BatchTopK SAE~\citep{bussmann2024batchtopk} per
(stream, suite, layer) at active budget $k{=}64$ on the post-block
residual streams of OpenVLA, the $\pi_{0.5}$ PaliGemma
vision-language backbone (PG), and the $\pi_{0.5}$ action expert (AE).

Table~\ref{tab:setup}
reports SAE sizes, offline diagnostics (FVE, alive fraction, average
$L_0$), and Hooked SR (10 rollouts per task) per (stream, layer).
Offline diagnostics and Hooked SR remain stable across SAE training seeds
(Appendix~\ref{app:ranking-stability}).
Full hyperparameters appear in Appendix~\ref{app:sae-training},
and sanity checks on hook location and information retention in
Appendix~\ref{app:sanity-checks}.

\begin{table}[!ht]
\centering
\scriptsize
\setlength{\tabcolsep}{2.5pt}
\begin{tabular}{llcccccc}
\toprule
Policy stream & Layer & SAE size & FVE $\uparrow$ & Alive\% & Avg $L_0$ & Raw SR (\%) & Hooked SR (\%) \\
\midrule
\textit{OpenVLA residual} & 0  & $4096{\rightarrow}32768$ & 0.911 & 94.1\% & 63.6 & $68.0{\pm}14.3\%$ & $0.0{\pm}0.0\%$ \\
& 16 & $4096{\rightarrow}32768$ & 0.999 & 3.1\%  & 63.4 &  & $0.0{\pm}0.0\%$ \\
& 24 & $4096{\rightarrow}32768$ & 0.998 & 3.4\%  & 64.3 &  & $0.2{\pm}0.5\%$ \\
& 31 & $4096{\rightarrow}32768$ & 0.995 & 8.0\%  & 63.8 &  & $34.8{\pm}24.5\%$ \\
\midrule
\textit{$\pi_{0.5}$ PG backbone} & 0  & $2048{\rightarrow}2048$ & 0.982 & 99.9\%  & 63.2 & $96.8{\pm}2.7\%$ & $96.5{\pm}4.0\%$ \\
& 5  & $2048{\rightarrow}2048$ & 0.987 & 100.0\% & 63.0 &  & $96.5{\pm}6.4\%$ \\
& 11 & $2048{\rightarrow}2048$ & 0.963 & 100.0\% & 63.0 &  & $97.8{\pm}3.9\%$ \\
& 16 & $2048{\rightarrow}2048$ & 0.996 & 99.9\%  & 64.3 &  & $95.8{\pm}2.2\%$ \\
\midrule
\textit{$\pi_{0.5}$ action expert} & 0  & $1024{\rightarrow}1024$ & 0.997 & 40.6\% & 64.0 & $96.4{\pm}2.6\%$ & $96.5{\pm}2.5\%$ \\
& 5  & $1024{\rightarrow}1024$ & 0.997 & 14.8\% & 64.0 &  & $95.8{\pm}3.9\%$ \\
& 11 & $1024{\rightarrow}1024$ & 0.984 & 84.9\% & 64.0 &  & $95.8{\pm}2.2\%$ \\
& 17 & $1024{\rightarrow}1024$ & 0.959 & 96.7\% & 63.9 &  & $97.2{\pm}2.9\%$ \\
\bottomrule
\end{tabular}
\caption{
This table reports suite-averaged offline SAE diagnostics and closed-loop
fidelity for each policy stream and layer, with means $\pm$ standard
deviations taken across the four LIBERO suites. Raw SR is the no-hook
policy success rate. Hooked SR is the success rate when the SAE
reconstruction replaces the hidden state at the corresponding layer.
PG layer 17 is omitted because the final PG prefix output is discarded
after KV-cache construction.
}
\label{tab:setup}
\end{table}

\textbf{SAE quality and fidelity differ sharply across VLA architectures.}
Alive fractions vary widely across model and layer
(Table~\ref{tab:setup}): OpenVLA drops from $94\%$ at layer~0 to
$3$--$8\%$ at deeper layers, $\pi_{0.5}$ PG saturates near $100\%$
throughout, and AE ranges from $14\%$ to $97\%$. Despite this
heterogeneity, all SAEs reach high reconstruction
(FVE $\geq 0.91$, mostly $\geq 0.96$). High reconstruction quality,
however, does not necessarily preserve closed-loop performance: OpenVLA Hooked SR is near zero at
layers 0, 16, and 24 and only $34.8\%$ at layer~31 (vs.\ a $68.0\%$ baseline), while $\pi_{0.5}$ sustains
$\geq 95\%$ Hooked SR at every probed PG and AE layer. Neither FVE
nor alive fraction alone predicts Hooked SR, so we use Hooked SR to assess behavioral fidelity.

\textbf{Why early-layer OpenVLA hooks fail.}
OpenVLA decodes each action as seven tokens over $256$ quantization
bins, so small reconstruction errors at early layers can shift
selected bins and compound across closed-loop steps. Failed rollouts
typically collapse to near-zero actions and stay there, since the
policy receives no observation change to drive it out of this
self-reinforcing stall. $\pi_{0.5}$ avoids this failure: its continuous flow-matching action
chunks are more robust to small reconstruction perturbations than
discrete-token outputs, consistent with \citet{grant2026not}'s
observations on action-token structure.
Layer~31 sits closest to the OpenVLA action head and leaves least
room for amplification, so we use it for all subsequent OpenVLA
discovery and intervention experiments.

\subsection{Keyframe Extraction and Event Clustering}
\label{sec:exp-event-clustering}

We extract keyframes and cluster them within each task using the methods
in Sections~\ref{sec:keyframe-extraction} and~\ref{sec:clustering}.
Table~\ref{tab:event-stats} summarizes the resulting keyframe and event-cluster statistics.
Most raw event types survive the $50\%$ episode-coverage filter as recurring
clusters, and counts are comparable across the two policies on each
suite, indicating similar end-effector kinematics despite the
different action interfaces.

\begin{table}[!ht]
\centering\small
\setlength{\tabcolsep}{6pt}
\begin{tabular}{llrrr}
\toprule
Source & Suite & Keyframes/rollout & Clusters & Recurring \\
\midrule
OpenVLA & LIBERO-Spatial & 4.15 & 48 & 36 \\
OpenVLA & LIBERO-Object  & 4.25 & 58 & 39 \\
OpenVLA & LIBERO-Goal    & 3.86 & 50 & 37 \\
OpenVLA & LIBERO-10      & 6.14 & 74 & 61 \\
\midrule
$\pi_{0.5}$ & LIBERO-Spatial & 4.11 & 46 & 36 \\
$\pi_{0.5}$ & LIBERO-Object  & 4.75 & 47 & 45 \\
$\pi_{0.5}$ & LIBERO-Goal    & 3.88 & 44 & 35 \\
$\pi_{0.5}$ & LIBERO-10      & 6.91 & 63 & 56 \\
\bottomrule
\end{tabular}
\caption{Keyframe and event-cluster statistics.
\emph{Keyframes/rollout}: AWE-proposed kinematic keyframes per
rollout. \emph{Clusters}: total task-local event clusters per suite.
\emph{Recurring}: clusters that pass the $50\%$ episode-coverage filter and enter the
suite-level rankings (\S\ref{sec:feature-ranking}).}
\label{tab:event-stats}
\end{table}

Recurring clusters span manipulation phases (approach, contact/grasp,
transport, release, withdrawal) and enter the event-aligned and window-mean rankings in
Section~\ref{sec:feature-ranking}. VLM-assigned phrase and phase
labels make each cluster human-readable:
Figure~\ref{fig:event-cluster-examples} shows per-cluster keyframe
bundles, and Figure~\ref{fig:event-feature-matrix} overlays the phase
labels onto the rows of the event-feature score heatmaps. Together
these views show whether a feature scores highly for one specific event (a single row) or across multiple phases.

\subsection{Closed-Loop Causal Intervention and Validation}
\label{sec:exp-causal}

Closed-loop intervention experiments test which SAE features
actually drive policy behavior. We focus on \emph{single-feature}
interventions throughout this section, since multi-feature edits
entangle effects across features and obscure attribution. We
compare two intervention modes: \emph{hard} zero-out ($\alpha{=}0$),
which silences a feature entirely and provides a clean on/off readout
of feature causality; and \emph{soft} intervention
($\alpha \in [0,1]$), which sweeps edit strength continuously and is
motivated by the saturation we observe under hard zero-out. 

We begin with hard zero-out, the strongest single-feature intervention
available under our protocol. For each (policy, layer, suite), we
compare four feature ranking strategies from
Section~\ref{sec:feature-ranking}:
\textsc{event-aligned}, \textsc{window-mean}, \textsc{task-mean}, and a
\textsc{random-alive} ranking that uniformly samples from alive features
after excluding all top selections of the other three rankings. For each ranking we
take the top $K$ features ($K{=}5$ for OpenVLA, $K{=}3$ for
$\pi_{0.5}$) and apply the residual-preserving latent edit of
Section~\ref{sec:intervention} to each feature separately at edit
strength $\alpha=0$. We report
$\Delta\mathrm{SR} = \mathrm{SR}_{\alpha=0} - \mathrm{SR}_{\mathrm{baseline}}$
(negative values indicate a drop), averaged across the $K$ features.
Rollout budgets are detailed in Appendix~\ref{app:rollout-budgets}. A ranking is
behaviorally informative when its top features produce more negative
$\Delta\mathrm{SR}$ than the \textsc{random-alive} ranking.
We also compare SAE interventions with norm-matched interventions along
non-SAE directions (Appendix~\ref{app:steering-safety-diagnostics}).

\begin{table*}[!ht]
\centering
\scriptsize
\setlength{\tabcolsep}{5pt}
\renewcommand{\arraystretch}{1.08}
\begin{tabular}{llccccc}
\toprule
Policy stream & Layer & Baseline SR (\%)
& \multicolumn{4}{c}{Zero-out SR (\%) with $\Delta$SR in parentheses} \\
\cmidrule(lr){4-7}
& & & Event-aligned & Window-mean & Task-mean & Random-alive \\
\midrule
OpenVLA residual & 31
& 68.0\%
& 48.8\% (-19.2)
& 63.8\% (-4.2)
& 63.5\% (-4.5)
& 68.7\% (+0.7) \\
\midrule
$\pi_{0.5}$ PG backbone & 0
& 96.8\%
& 94.7\% (-2.2)
& 95.5\% (-1.4)
& 96.0\% (-0.9)
& 96.2\% (-0.7) \\
& 5
& 96.8\%
& 96.0\% (-0.9)
& 95.8\% (-1.0)
& 96.7\% (-0.2)
& 96.3\% (-0.5) \\
& 11
& 96.8\%
& 94.2\% (-2.7)
& 94.7\% (-2.2)
& 94.0\% (-2.8)
& 95.8\% (-1.0) \\
& 16
& 96.7\%
& 96.0\% (-0.7)
& 96.7\% (+0.0)
& 95.2\% (-1.5)
& 96.3\% (-0.3) \\
\midrule
$\pi_{0.5}$ action expert & 0
& 96.4\%
& 0.0\% (-96.4)
& 0.0\% (-96.4)
& 0.0\% (-96.4)
& 95.7\% (-0.7) \\
& 5
& 96.4\%
& 0.8\% (-95.6)
& 0.2\% (-96.2)
& 0.2\% (-96.2)
& 73.0\% (-23.4) \\
& 11
& 96.4\%
& 0.0\% (-96.4)
& 0.2\% (-96.2)
& 0.0\% (-96.4)
& 88.3\% (-8.1) \\
& 17
& 96.4\%
& 15.2\% (-81.2)
& 0.0\% (-96.4)
& 0.0\% (-96.4)
& 89.3\% (-7.1) \\
\bottomrule
\end{tabular}
\caption{Closed-loop SR after zeroing selected SAE features. Mean SR
(\%) is followed in parentheses by
$\Delta\mathrm{SR}=\mathrm{SR}_{\alpha=0}-\mathrm{SR}_{\mathrm{baseline}}$
in percentage points (negative indicates lower success). Each entry
averages over the top $K$ features per ranking.}
\label{tab:causal-results-summary}
\end{table*}

\noindent\textbf{Cross-architecture results.}
Table~\ref{tab:causal-results-summary} shows three distinct
intervention regimes. For OpenVLA layer~31, the event-aligned
ranking produces the largest SR drop, well separated from the
random-alive ranking, while window-mean and task-mean rankings fall
in between. Temporal event alignment thus adds useful information
beyond feature activation magnitude alone. Within the OpenVLA
event-aligned ranking, the top three features carry nearly all of the
SR drop (per-feature breakdown in
Appendix~\ref{app:pi05-topk}), suggesting that a small set of high-weight features
dominates the closed-loop signal. 

The $\pi_{0.5}$ PaliGemma backbone stream behaves very differently:
across all probed layers and ranking strategies, single-feature
edits leave SR close to baseline, suggesting limited direct
leverage under the residual-preserving zero-out protocol.
These results suggest that modifying multiple features may be needed to affect PG
behavior (see Appendix~\ref{app:joint-dropout} for a joint zero-out probe).
By contrast, the action-expert stream collapses under the same
protocol, with top-ranked AE features driving SR to near zero across
most (layer, ranking) combinations. This collapse is specific to
the scored rankings: random-alive features reduce SR by only 0.7--23.4 points,
against 81.2--96.4 points for the three score-based rankings
(Table~\ref{tab:causal-results-summary}).
Qualitative failure examples show that the SR drops correspond to visible
closed-loop degradation rather than only minor trajectory deviations
(Appendix~\ref{app:zeroout-failure-examples}).
We read the AE results as evidence of broad causal sensitivity
rather than event-specific selectivity.
Neither $\pi_{0.5}$
stream shows comparable top-$K$ concentration: PG drops stay flat
across ranks, and AE drops are uniformly deep across top-1/2/3.
This is consistent with the architectural roles: the PaliGemma
backbone carries vision-language information that reaches the action
expert only indirectly through the KV-cache, while AE features
directly modulate the residual stream that produces the action chunk.

For both OpenVLA and the $\pi_{0.5}$ action expert, a target/off-target probe
on event-selected features finds nearly identical SR drops
(Appendix~\ref{app:target-offtarget-zeroout}), suggesting many
top-ranked features participate in shared manipulation computations
across tasks within a suite.

\begin{table}[!ht]
\centering\small
\setlength{\tabcolsep}{6pt}
\begin{tabular}{lccc}
\toprule
Setting & Event $\cap$ Window & Event $\cap$ Task & Window $\cap$ Task \\
\midrule
$\pi_{0.5}$ all       & $48.0\%$ & $48.0\%$ & $97.0\%$ \\
$\pi_{0.5}$ AE        & $43.7\%$ & $43.7\%$ & $93.7\%$ \\
$\pi_{0.5}$ PG        & $52.0\%$ & $52.0\%$ & $100.0\%$ \\
OpenVLA layer~31      & $5.0\%$  & $5.0\%$  & $95.0\%$ \\
\bottomrule
\end{tabular}
\caption{This table reports average top-$K$ feature overlap as a
percentage of $K$ between ranking strategies. \textsc{random-alive}
overlap with the other three rankings is zero by construction.
Window-mean and task-mean nearly coincide; event-aligned selects
largely distinct features.}
\label{tab:ranking-overlap}
\end{table}

\begin{figure}[!ht]
\centering
\begin{subfigure}[t]{0.421818\linewidth}
\centering
\includegraphics[width=\linewidth]{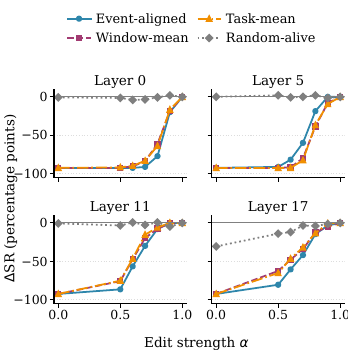}
\caption{LIBERO-10 suppression.}
\label{fig:ae-libero10-alpha-sweep}
\end{subfigure}\hfill
\begin{subfigure}[t]{0.56\linewidth}
\centering
\includegraphics[width=\linewidth]{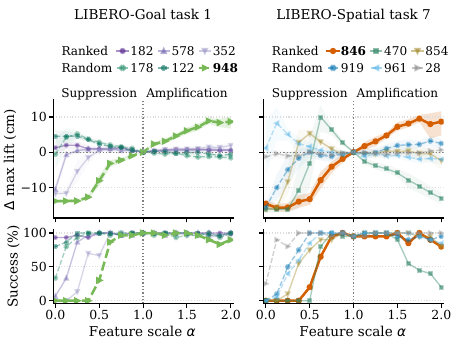}
\caption{Bowl-lift scaling at layer 5.}
\label{fig:bowl-lift-search}
\end{subfigure}
\caption{Suppressing and amplifying $\pi_{0.5}$ action-expert features.
(a) Change in success rate on LIBERO-10 at four layers, one curve per ranking.
(b) Feature scaling on LIBERO-Goal task 1 (left) and LIBERO-Spatial task 7 (right),
each with three ranked and three randomly sampled features. The upper plots show
changes in maximum bowl lift relative to the same-state no-edit rollout, including
failures; the lower plots show task success. Bold curves highlight features whose
amplification increases bowl lift. Shading indicates 95\% confidence intervals.}
\label{fig:feature-scaling}
\end{figure}

\noindent\textbf{Soft intervention: $\alpha$ as a controllable dial.}
The hard zero-out failures and the AE saturation in
Table~\ref{tab:causal-results-summary} both arise from a single
hyperparameter: the edit strength $\alpha=0$. To trace how behavior
changes with intervention magnitude, we sweep $\alpha$ continuously
between $0$ (full zero-out) and $1$ (identity).
Figure~\ref{fig:ae-libero10-alpha-sweep} shows that for $\pi_{0.5}$ AE
on LIBERO-10, SR transitions monotonically between collapse and
baseline at every probed layer, so adjusting $\alpha$ allows intermediate levels of disruption
rather than only the two extremes of no edit and collapse. The transition is steep, however: SR
rises rapidly over a narrow band of $\alpha$, so binary SR responds
more like a switch than a gradual dial, without describing how the robot's motion changes. This makes real-robot tuning
sensitive to small mis-calibrations: a slightly too-aggressive
$\alpha$ can push the policy into the arm-instability failure mode
visualized in Figure~\ref{fig:zeroout-failure-cases}.
Beyond intervention strength, we examine intervention timing and duration in
Appendix~\ref{app:windowed-suppression}.

\noindent\textbf{Beyond binary success: adjusting object lift.}
To examine behavioral changes not captured by task success alone, we
measure maximum bowl lift on two LIBERO bowl-placement tasks, both of which
require lifting the bowl before placement. This complements prior work
measuring end-effector height under VLA steering~\citep{haon2025mechanistic}. We scale
ranked and randomly sampled features in the $\pi_{0.5}$ action expert at
layer 5 over $\alpha\in[0,2]$. Among the screened features, amplifying
randomly sampled feature 948 on LIBERO-Goal and ranking-selected feature
846 on LIBERO-Spatial increases bowl lift across multiple scales
(Figure~\ref{fig:bowl-lift-search}). At $\alpha=1.5$, mean lift increases
by 6.1~cm and 7.2~cm relative to no intervention, with 30/30 and 20/20
successful trials, respectively. Beyond the success--failure transition above, increasing lift height while preserving task
success shows that feature edits can change how the robot performs a task.
Inducing a new behavior with a
single-feature edit is harder, since successful execution requires coordinating
multiple actions under changing observations
(Appendix~\ref{app:positive-steering}).

\noindent\textbf{Case study: ranking overlap.}
We also characterize what each ranking selects
(Table~\ref{tab:ranking-overlap}). Window-mean and task-mean rankings
nearly coincide: features highly active in event windows tend to be
highly active throughout the task. Event-aligned rankings, in contrast,
select largely distinct features, showing that temporal change does not
require high baseline magnitude. The same ranking strategies also select different feature sets across
architectures: event-aligned and magnitude rankings are almost
disjoint on OpenVLA layer~31 but overlap by $\sim$50\% on
$\pi_{0.5}$.
Appendix~\ref{app:ranking-stability} examines how stable these selections
are across rollout samples and SAE training seeds.

\subsection{Real-Robot Demonstration and Steering Safety}
\label{sec:exp-real-robot}

\noindent\textbf{Setup.}
A single-arm Mobile ALOHA platform with a wrist-mounted and a
third-person camera runs a LoRA-finetuned~\citep{hu2022lora}
$\pi_{0.5}$~\citep{intelligence2025pi05} on a chip-approach task. Each
rollout is conditioned on either \texttt{``Approach the red chips''} or
\texttt{``Approach the yellow chips''} with counterbalanced layouts so
successful behavior requires visual localization rather than trajectory
replay. We analyze SAE features in the $\pi_{0.5}$ action expert, using
SAEs trained on real-robot rollouts. Appendix~\ref{app:real-robot-setup}
gives the hardware details and a baseline reproduction of
\citet{haon2025mechanistic}'s FFN value-vector projection.

\begin{figure}[!ht]
\centering
\begin{minipage}[c]{0.25\linewidth}
\centering
\includegraphics[width=\linewidth]{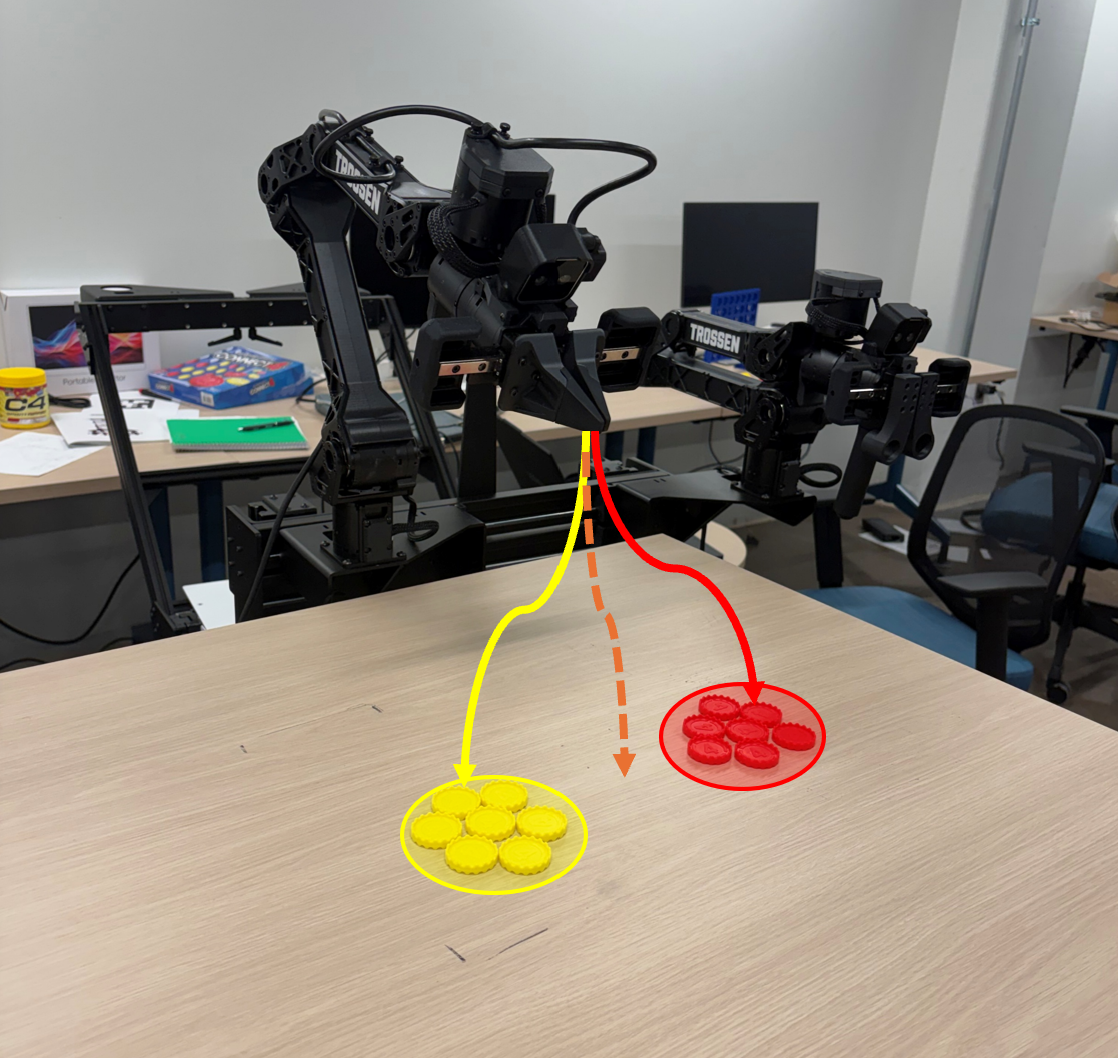}
\end{minipage}\hfill
\begin{minipage}[c]{0.73\linewidth}
\centering
\includegraphics[width=\linewidth]{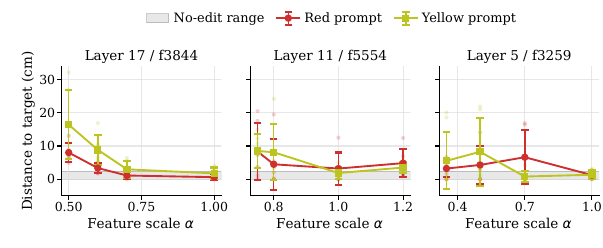}
\end{minipage}
\caption{Real-robot steering. Left: The chip-approach setup with a perturbed
trajectory (orange dashed arrow). Right: End-effector distance to the prompted
chip cluster versus SAE feature scale at three action-expert layers.
Lower distances indicate more faithful prompt-following. Error bars show
standard deviations, faint dots show individual trials, and the gray band
shows the range of distances observed without intervention.}
\label{fig:real-robot-steering-strength}
\end{figure}

\noindent\textbf{Steering effects depend on layer and strength.}
For all hardware trials, we restrict the tested range of SAE feature scales because aggressive or poorly tuned interventions can collapse behavior in simulation (Figure~\ref{fig:ae-libero10-alpha-sweep}) and induce large real-robot motions, posing risks to the robot and nearby personnel.
Figure~\ref{fig:real-robot-steering-strength} shows SAE scaling
measurably alters color-conditioned approach behavior, with the
strongest layer-dependent effect at action-expert layer~17 (stronger
suppression moves the end-effector farther from the prompted cluster);
earlier layers are noisier. Selected SAE features can therefore
affect real-robot prompt-following, with effect magnitude depending on
layer, feature, and feature scale. These interventions probe causal sensitivity. Achieving reliable control
over which color target the robot approaches and establishing safe limits on SAE feature
scales for physical deployment remain future work.

\section{Limitations and Conclusion}
\label{sec:limitations}

\textbf{Limitations.}
(i) Finer-grained feature steering: we identify features that influence execution and can
scale them to adjust motion. However, these features may represent motion
patterns shared across tasks, making it difficult to isolate individual
behavioral variables. Choosing a feature and a strength for a desired change
remains an open problem.
(ii) Feature semantics: VLM labels describe event clusters, but the semantic
meaning of individual SAE features remains unclear. These labels are used
mainly for visualization, while their use in feature ranking and semantic validation
remains unexplored.
(iii) Behavioral evaluation: binary SR does not reveal how behavior changes.
Lift height and approach distance provide finer readouts,
but broader benchmarks are needed to test whether edits reliably achieve
specified motion changes safely across scenes. The cost of closed-loop
rollouts limits the coverage of these tests.

\textbf{Conclusion.} We introduce event-grounded SAE analysis as a
behavior-anchored approach to VLA mechanistic interpretability. By
linking kinematic keyframes, task-local event clusters, and
event-aligned ranking, the pipeline scores every alive SAE feature
without per-feature labeling and selects candidates for closed-loop
causal testing. Across OpenVLA and $\pi_{0.5}$ in simulation, and
on a real-robot LoRA-finetuned $\pi_{0.5}$, it reveals
architecture- and site-dependent
intervention effects. Scaling selected SAE features can adjust lift height
in simulation and approach distance on a physical robot. Event-grounded SAE
analysis, anchored in behavioral events defined independently of SAE activations, provides a starting point for studying and adjusting VLA
behavior. Future work could extend event proposals beyond end-effector motion,
move VLM-derived
event semantics from visualization into ranking and intervention,
replace binary success with finer-grained behavioral metrics, and
develop safety-calibrated interventions for physical deployment. Together, these steps move toward more transparent and controllable VLA policies for real-world robotics.

\begin{ack}
We have no external funding or competing interests to disclose.
\end{ack}

\newpage
\bibliographystyle{unsrtnat}
\bibliography{reference}

\newpage
\appendix

\FloatBarrier
\section{SAE Training Details}
\label{app:sae-training}

We train one BatchTopK SAE per policy stream, LIBERO suite, and layer using
closed-loop activation collections from 50 rollouts per task. OpenVLA examples
are residual states from autoregressive action-token forwards. PG examples are
residual states from the PaliGemma prefix-prefill pass. AE examples are residual
states from action-expert denoising forwards over predicted action chunks.
We use the training settings in Table~\ref{tab:hparams}\subref*{tab:hparams-sae},
except for the OpenVLA SAEs on LIBERO-Spatial at layers 0, 16, and 31, which use
a learning rate of $5\times10^{-5}$, 4{,}000 training steps, and a batch size
of 4{,}096.

\paragraph{Hook locations.}
All SAEs are trained and applied at post-block residual states. For OpenVLA, the
hook is placed after each selected language-model transformer block during
action-token decoding. For PG, the hook is placed after each selected PaliGemma
prefix block during the prefix-prefill pass. PG layer 17 is excluded because the
final prefix hidden state is discarded after KV-cache construction, so a
post-block intervention at PG layer 17 cannot affect the action expert. For AE,
the hook is placed after each selected action-expert block during denoising.
AE activations are indexed by the policy query timestep, denoising step, and
action-token offset. Action-token offset $j$ is aligned to environment timestep
$t+j$.

\begin{table}[!ht]
\centering
\small
\begin{subtable}[t]{0.52\linewidth}
\centering
\setlength{\tabcolsep}{4pt}
\begin{tabular}{lc}
\toprule
Hyperparameter & Value \\
\midrule
OpenVLA layers & $\{0,16,24,31\}$ \\
PG layers & $\{0,5,11,16\}$ \\
AE layers & $\{0,5,11,17\}$ \\
OpenVLA SAE size & $4096 \rightarrow 32768$ \\
PG SAE size & $2048 \rightarrow 2048$ \\
AE SAE size & $1024 \rightarrow 1024$ \\
Active budget $k$ & 64 \\
Learning rate & $1\times 10^{-4}$ \\
Batch size & 40{,}000 \\
Optimizer & Adam \\
Training steps & 15k OpenVLA; 10k PG/AE \\
Warmup steps & 1{,}000 \\
Decay start & 80\% of training \\
Activation normalization & Enabled \\
Training dtype & float32 \\
\bottomrule
\end{tabular}
\subcaption{SAE training hyperparameters.}
\label{tab:hparams-sae}
\end{subtable}\hfill
\begin{subtable}[t]{0.46\linewidth}
\centering
\setlength{\tabcolsep}{4pt}
\begin{tabular}{lc}
\toprule
Hyperparameter & Value \\
\midrule
AWE error budget $\eta$ & $0.05$ \\
Scoring window radius $w$ & $5$ \\
\midrule
Visual weight $\lambda_v$ & $1.0$ \\
State weight $\lambda_s$ & $0.5$ \\
Temporal weight $\lambda_p$ & $0.4$ \\
\midrule
Cosine threshold (agglomerative) & $0.18$ \\
Cluster coverage threshold & $0.5$ \\
\bottomrule
\end{tabular}
\subcaption{Clustering hyperparameters.}
\label{tab:hparams-clustering}
\end{subtable}
\caption{Reproducibility hyperparameters. (a) SAE training (size denotes
activation dimension $\to$ dictionary size). (b) Keyframe extraction
and event clustering.}
\label{tab:hparams}
\end{table}

\paragraph{Rollout counts.}
\label{app:rollout-budgets}
For each task, we use 50 rollouts to collect training activations, 50 to
evaluate the unmodified $\pi_{0.5}$ policy, and 10 to evaluate the unmodified
OpenVLA policy. Reconstruction-only evaluations use 10 rollouts per task for
each SAE. Single-feature interventions use 5 rollouts per task for each
selected feature and intervention strength.
Experiments ran on NVIDIA A100 GPUs, with each SAE requiring a few GPU-hours
to train.

\noindent\textbf{PaliGemma attention-valid analysis.} We also evaluate
PaliGemma statistics and feature rankings using only attention-valid positions
from the two available image views and the prompt text, excluding the absent
camera block and text padding. The SAEs remain those trained on all collected
positions. We repeat the analysis across four LIBERO suites at layers 0, 5, 11,
and 16. Suite-averaged FVE ranges from 0.852 to 0.995 across these layers.
Using the resulting rankings, single-feature zero-out gives mean SRs of
94.5--96.3\% across layer--ranking combinations, averaged over suites and selected
features, compared with 96.8\% without intervention.
These results are consistent with the main-text finding that single-feature
zero-out leaves PaliGemma task success close to baseline.

\FloatBarrier
\section{Clustering Hyperparameters}
\label{app:clustering-hyperparams}

This appendix details the kinematic keyframe extraction
(Section~\ref{sec:keyframe-extraction}) and event clustering
(Section~\ref{sec:clustering}) stages. The hyperparameter values are
given in Table~\ref{tab:hparams}\subref*{tab:hparams-clustering}.

\paragraph{Hyperparameter selection.}
We calibrated the descriptor weights $\lambda_v, \lambda_s, \lambda_p$ so
that the recurring-cluster count per task stays on the order of the
typical AWE waypoint count per rollout, keeping clusters aligned with the
kinematic event scale of the underlying trajectories.
The visual descriptor averages the embeddings of the five strip frames
from a frozen SigLIP vision encoder~\citep{zhai2023sigmoid}
(\texttt{google/siglip-base-patch16-224}).
Within each task, we standardize the robot state (3-D end-effector
position and gripper command) and rollout progress, then normalize the visual
and state descriptors. Their weighted concatenation with rollout progress
is normalized before clustering. The
relative ordering $\lambda_v > \lambda_s > \lambda_p$ reflects the
intuition that visual scene context is the most discriminative signal,
end-effector state breaks ties between visually similar moments, and
temporal progress provides a weak ordering prior.

\FloatBarrier
\section{VLM Cluster Annotation Prompt}
\label{app:vlm-prompt}

This appendix gives an abridged version of the prompt template used by the VLM cluster
annotation step (Section~\ref{sec:clustering}). We use
\texttt{gemini-3.1-pro-preview} as the VLM. The template is filled per cluster with the task
instruction, the cluster identifier, the episode coverage, and the
number of clips and frames-per-clip drawn from the cluster's keyframe
bundles. The phase taxonomy is adapted from~\citet{chen2026robovlm}.

\begin{small}
\begin{verbatim}
You are labeling a recurring event in a robot manipulation task.

You will be shown <N> image sequences from different rollout
episodes of the same task. Each sequence contains <F> images in
chronological order. All sequences were clustered automatically
and are intended to depict the same recurring event type. Treat
each sequence as one short clip; the model input is therefore a
small batch of clips.

Task instruction: "<task description>"
Cluster id: <cluster id>
Episode coverage: <coverage fraction>

Your job is to assign one canonical phrase describing the shared
event across the sequences. Focus on the common event across clips,
not small differences between episodes. Prefer a dynamic event
description (e.g., releasing, approaching) over a static state
description. If the later frames make the event clearer, prioritize
them when choosing the phrase and phase.

Choose exactly one phase label from the closed set:
  pre_grasp, immobilization, contact, detach, post_grasp,
  transition.

Return JSON with exactly two keys:
  {
    "phrase": <short canonical event phrase>,
    "phase":  <one label from the set above>
  }

The top-level JSON value must be a single object, not a list.
Output exactly one short human-readable phrase. Do not include
any keys other than "phrase" and "phase", and do not include
markdown fences.
\end{verbatim}
\end{small}

\FloatBarrier
\section{Feature Ranking Score Definitions}
\label{app:ranking-formulas}

This appendix gives the per-episode and per-cluster score formulas
for the four ranking strategies described in
Section~\ref{sec:feature-ranking}.
The final scores for event-aligned, window-mean, and task-mean ranking
are $S_{\mathrm{event}}$, $S_{\mathrm{window}}$, and $S_{\mathrm{task}}$,
respectively.

\textbf{Notation.} A \emph{rollout} is one execution of the policy on a
task. A \emph{keyframe} is a timestep selected by AWE, with an \emph{event
window} covering $w$ steps on either side for measuring feature activations.
An \emph{event} is a retained event cluster from Section~\ref{sec:clustering},
that is, a group of similar keyframes from rollouts of the same task.
An event may include keyframes from several rollouts,
possibly several from the same rollout, so we average window scores first
within each rollout and then across rollouts. We describe the scores for one SAE feature at a time. Let
$\bar z_t$ be its activation at environment step $t$, averaged over the
encoded vectors aligned to that step (Appendix~\ref{app:sae-training}).

\textbf{Window scores.} Event-aligned ranking compares each event window
with three temporal templates over offsets $\delta\in\{-w,\ldots,w\}$, shown
in Equation~\ref{eq:ranking-templates}.
\begin{equation}
q_{\mathrm{pulse}}(\delta)=1-\frac{|\delta|}{w+1},
\quad
q_{\mathrm{up}}(\delta)=
\begin{cases}
-1, & \delta<0,\\
+1, & \delta\geq0,
\end{cases}
\quad
q_{\mathrm{down}}=-q_{\mathrm{up}}.
\label{eq:ranking-templates}
\end{equation}
Each template is mean-centered and normalized to unit $\ell_2$ norm,
giving $\tilde q$. Mean-centering ensures that a constant activation
contributes zero, separating temporal change from activation level. The score
of window $i$ under template $q$ retains only positive projections,
$s_i^q=\operatorname{ReLU}\big(\sum_{\delta=-w}^{w}\tilde q(\delta)\,\bar z_{t_i+\delta}\big)$.
Window-mean ranking instead uses the mean activation in the window,
$\mu_i=\frac{1}{2w+1}\sum_{\delta=-w}^{w}\bar z_{t_i+\delta}$.

\textbf{Score aggregation.} For a fixed event $e$ in rollout $r$, let $n$
be the number of its keyframe windows. Event-aligned ranking averages the
window scores separately for the pulse, step-up, and step-down templates, then
keeps the highest average, $c_r$. Window-mean ranking averages the window means
to obtain $m_r$. Equation~\ref{eq:ranking-rollout-aggregation} gives both
rollout-level scores.
\begin{equation}
c_r=\max_q\left(\frac{1}{n}\sum_{i=1}^{n}s_i^q\right),
\qquad
m_r=\frac{1}{n}\sum_{i=1}^{n}\mu_i.
\label{eq:ranking-rollout-aggregation}
\end{equation}
The same event may appear in several rollouts, each providing scores $c_r$
and $m_r$. Let $R_e$ be the number of rollouts containing event $e$, indexed
here by $r=1,\ldots,R_e$. The event-level scores $c_e$ and $m_e$ average over
these rollouts, so each rollout containing the event receives equal weight
(Equation~\ref{eq:ranking-event-aggregation}).
\begin{equation}
c_e=\frac{1}{R_e}\sum_{r=1}^{R_e}c_r,
\qquad
m_e=\frac{1}{R_e}\sum_{r=1}^{R_e}m_r.
\label{eq:ranking-event-aggregation}
\end{equation}
Finally, let $M$ be the number of retained events across the suite and
$N_e$ the total number of keyframes belonging to event $e$ across rollouts.
The suite-level score $S_{\mathrm{event}}$ gives each event equal weight,
whereas $S_{\mathrm{window}}$ weights each event by its keyframe count
(Equation~\ref{eq:ranking-suite-aggregation}).
\begin{equation}
S_{\mathrm{event}}=\frac{1}{M}\sum_{e=1}^{M}c_e,
\qquad
S_{\mathrm{window}}=
\frac{\sum_{e=1}^{M}N_e m_e}{\sum_{e=1}^{M}N_e}.
\label{eq:ranking-suite-aggregation}
\end{equation}

\textbf{Task-mean ranking.} Task-mean ranking does not use events. We pool
the timesteps from all rollouts in the suite and reindex them as
$t=1,\ldots,T$, where $T$ is their total number. The score in
Equation~\ref{eq:ranking-task-aggregation} gives every timestep equal weight.
\begin{equation}
S_{\mathrm{task}}=\frac{1}{T}\sum_{t=1}^{T}\bar z_t.
\label{eq:ranking-task-aggregation}
\end{equation}

For the three score-based rankings, we select the top-$K$ features by
suite-level score as the candidate set evaluated in the intervention
experiments. When selecting candidates for a single task, we compute the
scores using only that task's rollouts and events.

\textbf{Random-alive ranking.}
A uniform sample of $K$ features from the set of alive SAE features
in the suite, excluding any feature already selected by the other
three rankings.

\FloatBarrier
\section{Real-Robot Setup}
\label{app:real-robot-setup}

This appendix describes the hardware, sensing, deployment, and logging
configuration for the real-robot study in Section~\ref{sec:exp-real-robot}.

\subsection{Hardware, Sensing, and Deployment}

The Mobile ALOHA platform~\citep{fu2024mobile} controls a single follower
arm over a $13.5 \times 30$~in tabletop workspace
(Figure~\ref{fig:real-robot-steering-strength}, left). Two Intel RealSense cameras provide
RGB observations at $640 \times 480$ and 30~Hz: a wrist-mounted egocentric
view on the active arm, and a third-person view from roughly 60~in above
the workspace. Policy inputs are joint-position state, images resized to
$224 \times 224$, and the natural-language prompt.

We deploy the LoRA-finetuned $\pi_{0.5}$ via the OpenPI server/client
pattern: a websocket server hosts the policy, and the real-robot client
handles sensing and action execution at 30~Hz. Each rollout runs for up
to 300 control steps, with the client requesting a new action chunk
every 50 steps (matching the chunk horizon $H=50$). At episode start,
the arm is interpolated to the first predicted action over 5~s. Between
episodes it returns to a captured home pose.

Per-step logs include task and episode identifiers, prompt, executed
actions, gripper state, timestamps, and end-effector state (reconstructed
via forward kinematics). For interpretability analysis, we record
$\pi_{0.5}$ action-expert activations during inference and align them to
environment timesteps via
$t_{\mathrm{env}} = t_{\mathrm{chunk}} + j$ for $j \in \{0,\dots,H-1\}$.
The hardware SAEs are trained on these activations at layers 0, 5, 11,
and 17, with 8{,}192 features, $k=64$, learning rate $10^{-4}$, 15{,}000 steps,
and batch size 40{,}000.

\subsection{Value-Vector Steering Baseline}
\label{app:value-vector}

We evaluate the value-vector method of \citet{haon2025mechanistic} using the
same fine-tuned $\pi_{0.5}$ policy and Mobile ALOHA chip-approach setup as in
Section~\ref{sec:exp-real-robot}. For red-, yellow-, and random-vector
steering, we keep the instruction fixed as ``Approach the chips.''
In the prompt-only baselines, we instead use ``Approach the red chips.''
or ``Approach the yellow chips.'' and do not modify the policy's
internal activations.

We project rows of the $\pi_{0.5}$ PaliGemma backbone's feed-forward output
matrices, with low-rank adaptation weights merged, into token embedding
space and select six directions associated with each color, using the seed
words \texttt{red}, \texttt{crimson}, \texttt{scarlet}, and \texttt{reddish}
for red and \texttt{yellow}, \texttt{gold}, \texttt{golden}, \texttt{amber},
and \texttt{yellowish} for yellow. We compare these directions with six random
vectors using the same steering coefficient of 10. Each condition uses 10
rollouts with balanced chip layouts.

\begin{figure}[!ht]
\centering
\includegraphics[width=0.65\linewidth]{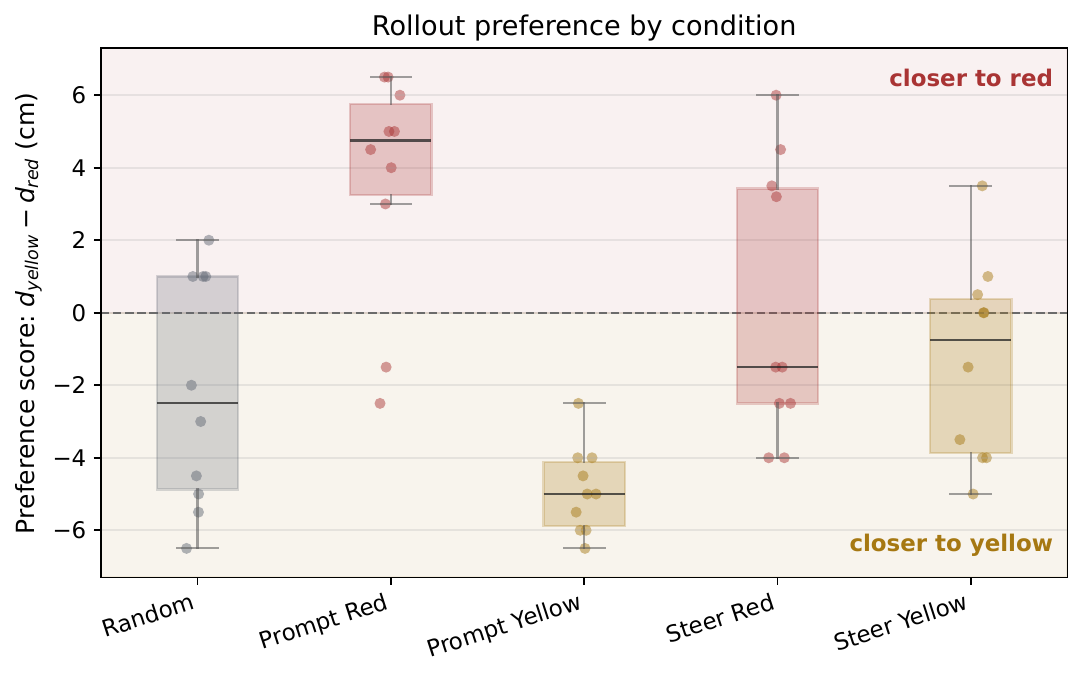}
\caption{Value-vector steering preference score
$\Delta d=d_{\mathrm{yellow}}-d_{\mathrm{red}}$, using minimum distances to the
two chip clusters. Positive values indicate closer approach to red and
negative values to yellow. Boxes show medians and quartiles across 10
rollouts, and dots mark individual rollouts.}
\label{fig:value-vector}
\end{figure}

Prompt-only baselines produce clearer color-conditioned behavior, while
value-vector steering is more variable and overlaps the random-vector
baseline (Figure~\ref{fig:value-vector}). In our experiments, value-vector
steering did not consistently shift the robot's motion toward the
corresponding color target.

\FloatBarrier
\clearpage
\section{Event-Aligned Feature Score Visualization}
\label{app:event-feature-matrix}

We examine whether high event-aligned scores are concentrated in a small
set of features shared across tasks, or whether different tasks have distinct
high-scoring features.

\begin{figure}[!htbp]
\centering
\includegraphics[width=\linewidth]{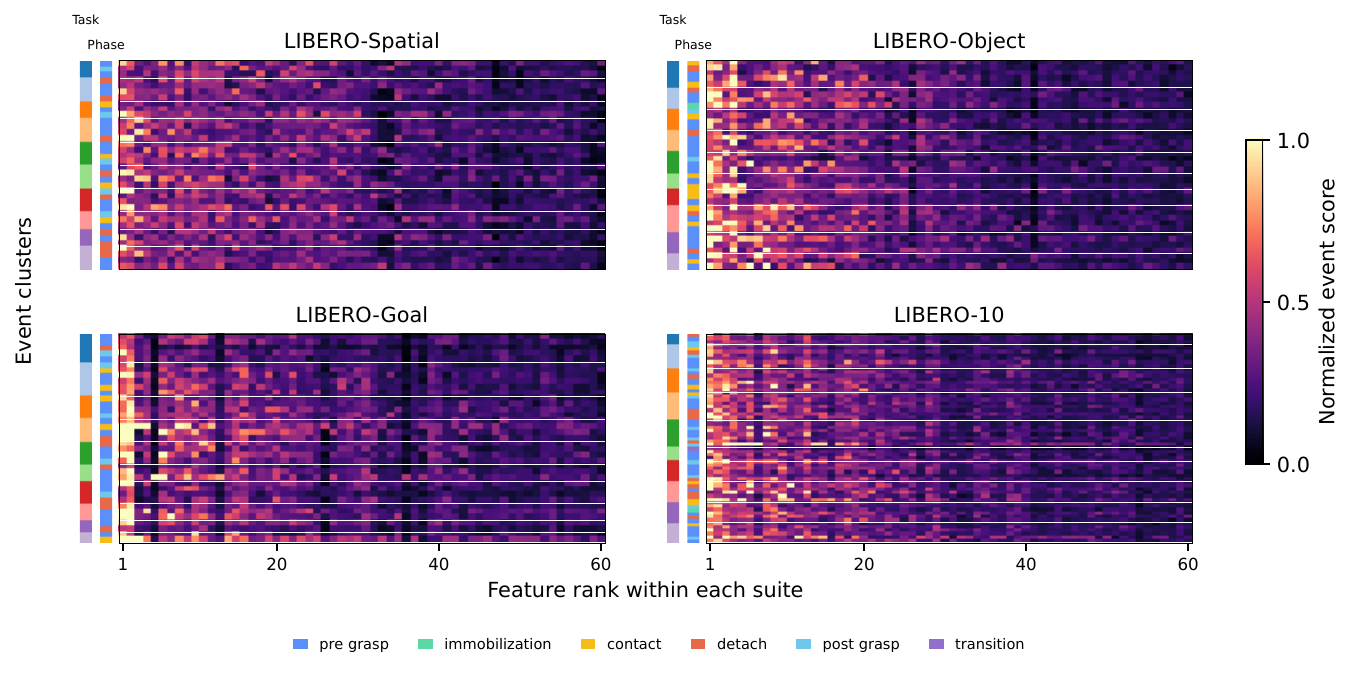}
\caption{Event-aligned score matrices for OpenVLA layer 31. Side strips
identify tasks and VLM-assigned manipulation phases.}
\label{fig:event-feature-matrix}
\end{figure}

\begin{figure}[!htbp]
\centering
\includegraphics[width=\linewidth]{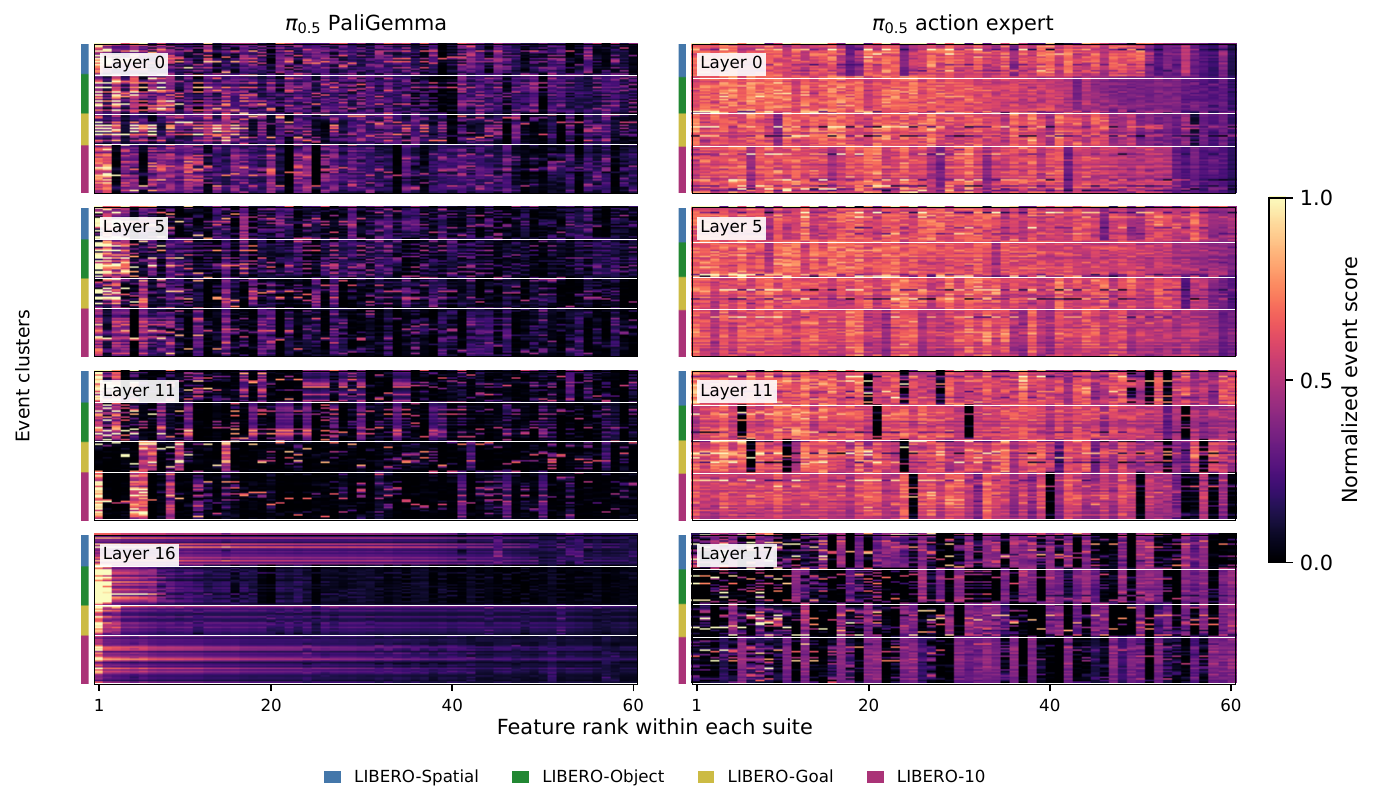}
\caption{Event-aligned score matrices for the $\pi_{0.5}$ PaliGemma backbone
(left) and action expert (right). Each matrix stacks events from four suites,
identified by side color strips. PaliGemma
scores use attention-valid prefix positions
(Appendix~\ref{app:sae-training}).}
\label{fig:pi05-event-feature-matrix}
\end{figure}

Figures~\ref{fig:event-feature-matrix} and~\ref{fig:pi05-event-feature-matrix}
show event-aligned score matrices across four LIBERO suites for OpenVLA layer
31, $\pi_{0.5}$ PaliGemma layers 0, 5, 11, and 16, and action-expert layers 0,
5, 11, and 17. Rows represent event clusters (Section~\ref{sec:clustering}),
and columns represent the top 60 features ranked by maximum event score
separately for each suite and SAE. Colors show event-aligned scores on a shared
scale across suites for each model, layer, and component. VLM phase annotations are provided only for
OpenVLA.

In OpenVLA and the $\pi_{0.5}$ PaliGemma backbone, high scores are
concentrated in a few features across events, whereas in the action expert
they are spread across many features. At action-expert layer 17, high-scoring
features vary more across events than at earlier layers.

\begin{figure}[!htbp]
\centering
\includegraphics[width=\linewidth]{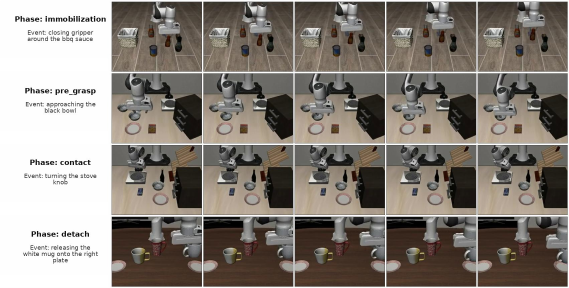}
\caption{This figure shows four representative OpenVLA task-local event clusters. Each
row shows one keyframe bundle as a five-frame strip at offsets
$\{-4,-2,0,2,4\}$ around the keyframe timestep, with the cluster's
VLM-assigned phrase and phase label.}
\label{fig:event-cluster-examples}
\end{figure}

One hypothesis is that event-related information is initially concentrated
in a few features, becomes more broadly distributed during action
computation, and is organized into more event-selective features near
the action output. Under this interpretation, early action-expert features
may participate in shared computations across events, whereas later
features may differentiate particular aspects of motion.
To examine whether this shared involvement also extends across tasks,
Appendix~\ref{app:target-offtarget-zeroout} compares intervention effects on the
task used for feature selection and a different task.
These heatmaps guide feature
discovery, with behavioral effects evaluated through closed-loop zero-out
experiments in Section~\ref{sec:exp-causal}.
Figure~\ref{fig:event-cluster-examples} shows representative keyframe bundles
from the event clusters.

\section{Zero-Out Failure Examples}
\label{app:zeroout-failure-examples}

\begin{figure}[!ht]
\centering
\begin{subfigure}{\linewidth}
    \centering
    \includegraphics[width=\linewidth]{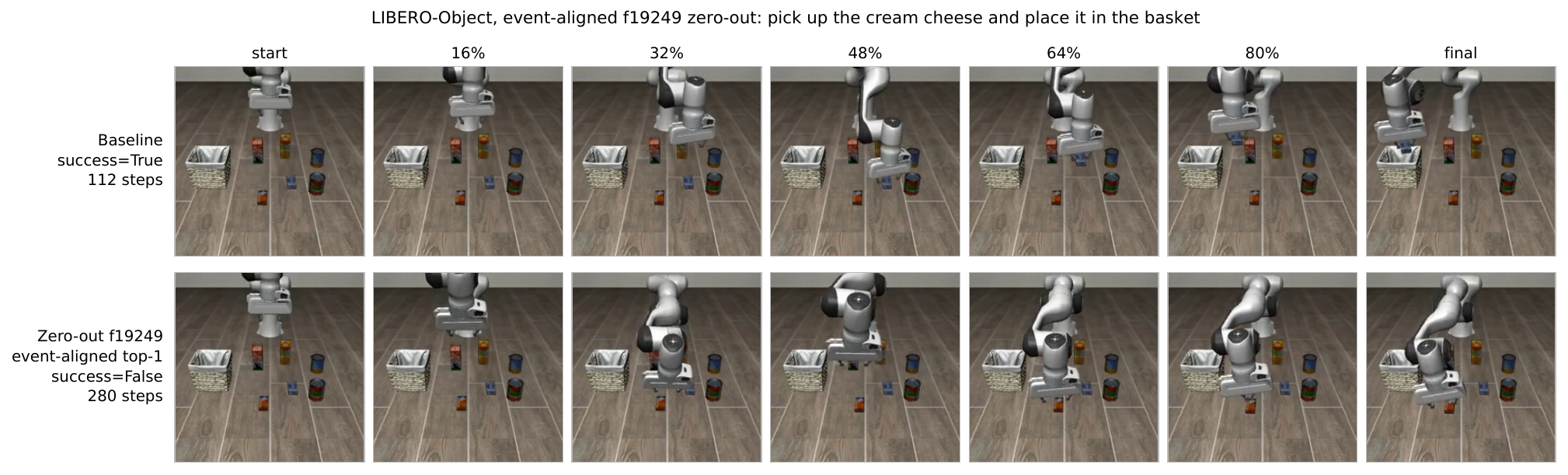}
    \caption{OpenVLA layer~31, event-aligned $f19249$ on
    LIBERO-Object \texttt{pick up the cream cheese and place it in the basket}: the
    baseline succeeds while zero-out fails target localization,
    grasping, and placement.}
    \label{fig:zeroout-failure-openvla}
\end{subfigure}

\begin{subfigure}{\linewidth}
    \centering
    \includegraphics[width=\linewidth]{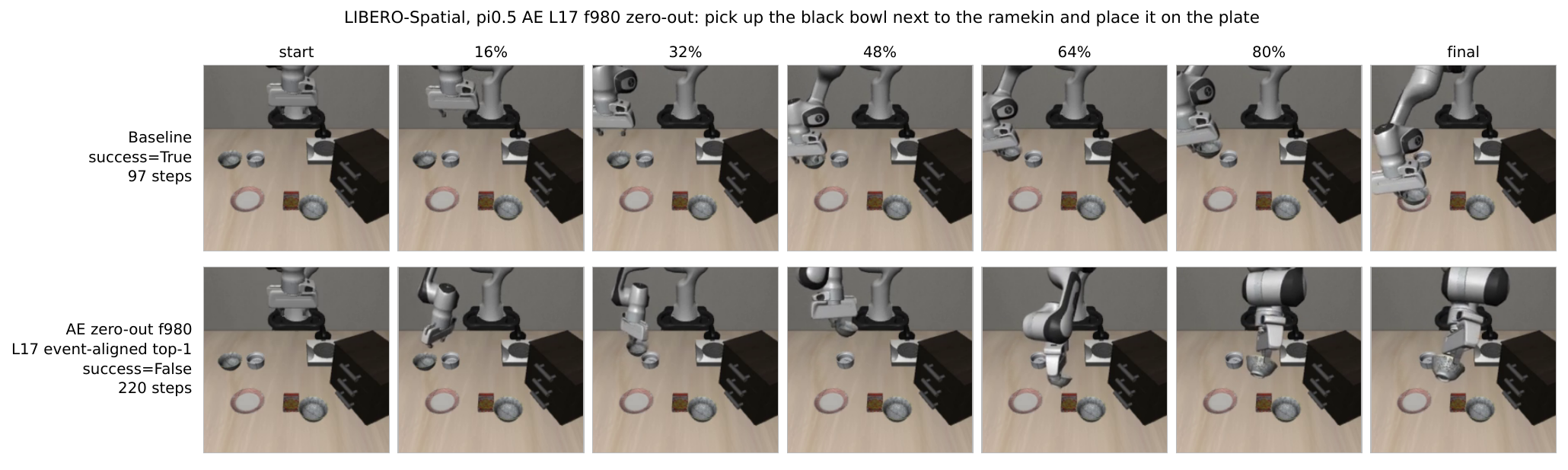}
    \caption{$\pi_{0.5}$ action-expert layer~17, event-aligned
    $f980$ on LIBERO-Spatial \texttt{pick up the black bowl next to the ramekin and place it on the plate}: the baseline completes in $97$ steps while zero-out
    destabilizes the arm and times out at $220$ steps.}
    \label{fig:zeroout-failure-pi05}
\end{subfigure}

\caption{Representative zero-out failures on both
VLA architectures show that the SR drops in
Table~\ref{tab:causal-results-summary} correspond to visible
closed-loop degradation rather than only minor trajectory deviations.}
\label{fig:zeroout-failure-cases}
\end{figure}

\begin{figure}[p]
\centering
\includegraphics[width=\linewidth]{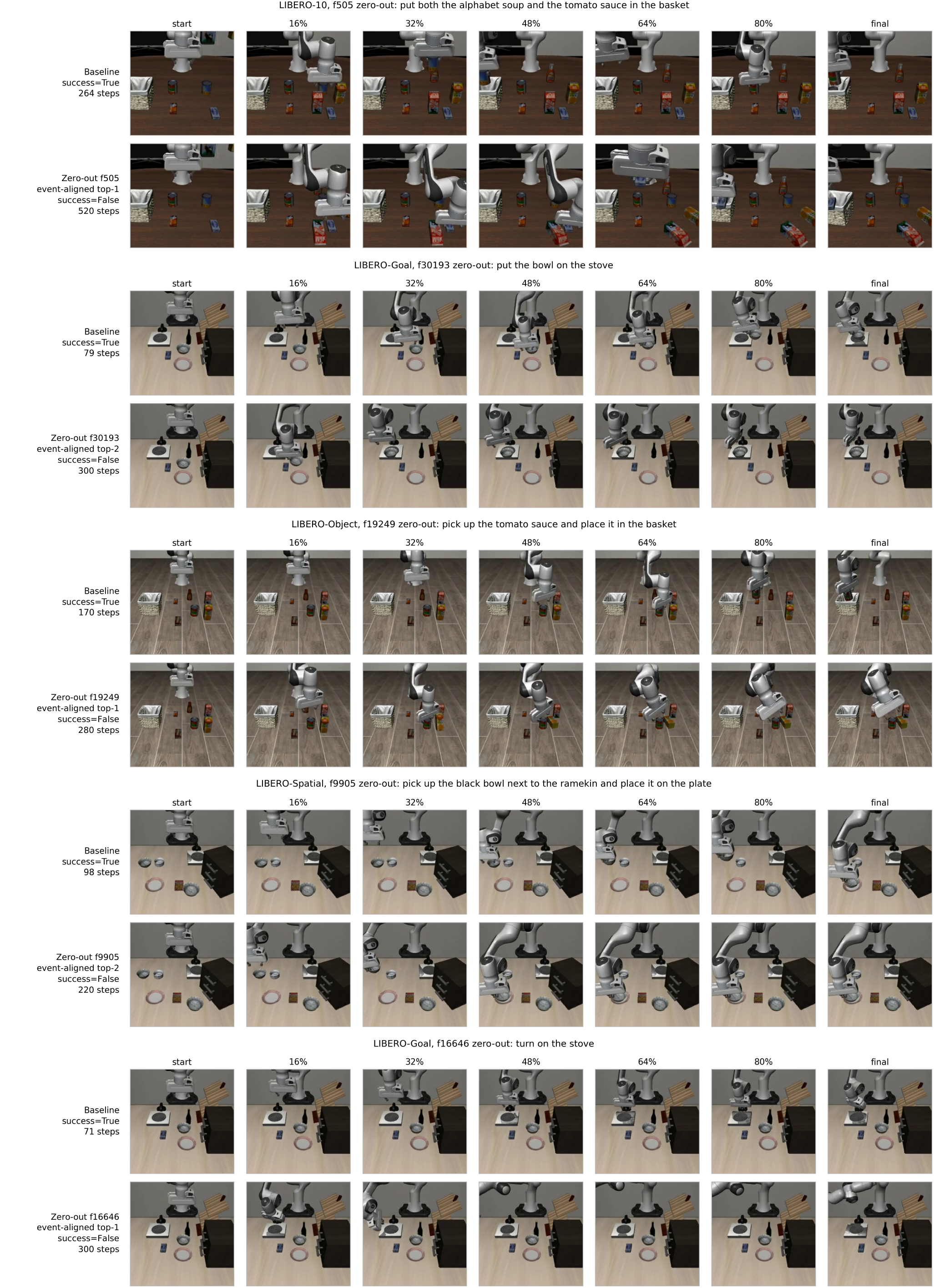}
\caption{Additional rollout snapshots compare baseline and single-feature zero-out runs.
Each pair shows the effect of ablating one event-score-selected OpenVLA layer-31 SAE feature.
Failures range from localized degradation (poor target localization,
mistimed gripper) to larger control changes (e.g.,
LIBERO-Goal \texttt{turn on the stove} with feature $f16646$ induces
markedly larger arm excursions).}
\label{fig:zeroout-failure-gallery}
\end{figure}

Figure~\ref{fig:zeroout-failure-cases} shows representative zero-out
failures on both architectures: the OpenVLA case disrupts target
localization, grasping, and placement, while the $\pi_{0.5}$
action-expert case destabilizes the arm trajectory until rollout
timeout. 
Figure~\ref{fig:zeroout-failure-gallery} reports an additional gallery
of paired baseline and zero-out rollout snapshots.

\section{Joint Dropout Probe (PaliGemma Backbone)}
\label{app:joint-dropout}

Joint zero-out of 16 random active PG features
(Table~\ref{tab:pg-active-drop16}) shows two patterns. (1) PG is
markedly more robust than AE: at layers 5, 11, and 16, even 16 features dropped at once give
moderate SR drops, far smaller than the near-total collapses single
AE features induce in Table~\ref{tab:causal-results-summary}. At layer 0, the same
dropout is severe, and LIBERO-Goal falls to 8\% SR. (2)
Within PG, the effect concentrates at early layers and weakens
with depth until the same intervention has essentially no effect at layer 16. This trend is consistent with
how
the action expert reads the prefix: a post-block edit at PG layer $\ell$
can affect only the keys and values of later prefix layers, so edits at
deeper layers reach fewer action-expert layers.

\begin{table*}[!ht]
\centering
\small
\setlength{\tabcolsep}{5pt}
\renewcommand{\arraystretch}{1.06}
\begin{tabular}{llccc}
\toprule
Suite & PG layer & Baseline SR (\%) & Dropout SR (\%) & $\Delta$SR (pp) \\
\midrule
LIBERO-Spatial & 0  & 98.0 & $66.7{\pm}6.8$  & $-31.3$ \\
               & 5  & 98.0 & $86.7{\pm}3.8$  & $-11.3$ \\
               & 11 & 98.0 & $98.0{\pm}1.6$  & $+0.0$  \\
               & 16 & 97.8 & $100.0{\pm}0.0$ & $+2.2$  \\
\midrule
LIBERO-Object  & 0  & 98.8 & $53.3{\pm}4.7$  & $-45.5$ \\
               & 5  & 98.8 & $88.7{\pm}7.7$  & $-10.1$ \\
               & 11 & 98.8 & $96.0{\pm}1.6$  & $-2.8$  \\
               & 16 & 98.8 & $100.0{\pm}0.0$ & $+1.2$  \\
\midrule
LIBERO-Goal    & 0  & 97.8 & $8.0{\pm}1.6$   & $-89.8$ \\
               & 5  & 97.8 & $79.3{\pm}2.5$  & $-18.5$ \\
               & 11 & 97.8 & $96.7{\pm}0.9$  & $-1.1$  \\
               & 16 & 96.8 & $96.0{\pm}1.6$  & $-0.8$  \\
\midrule
LIBERO-10      & 0  & 92.8 & $77.3{\pm}2.5$  & $-15.5$ \\
               & 5  & 92.8 & $84.0{\pm}3.3$  & $-8.8$  \\
               & 11 & 92.8 & $80.0{\pm}7.1$  & $-12.8$ \\
               & 16 & 93.2 & $94.7{\pm}0.9$  & $+1.5$  \\
\bottomrule
\end{tabular}
\caption{
This table reports closed-loop success after randomly dropping 16 active
PaliGemma SAE features. Values are percentages. Dropout SR reports the
mean success rate $\pm$ standard deviation across three random seeds.
$\Delta$SR is computed as
$\mathrm{SR}_{\mathrm{dropout}}-\mathrm{SR}_{\mathrm{baseline}}$ in percentage
points. Negative values indicate lower success after intervention.
}
\label{tab:pg-active-drop16}
\end{table*}
\FloatBarrier

\section{Per-Feature Top-$K$ Zero-Out Breakdown}
\label{app:pi05-topk}

\begin{figure}[!ht]
\centering
\includegraphics[width=\linewidth]{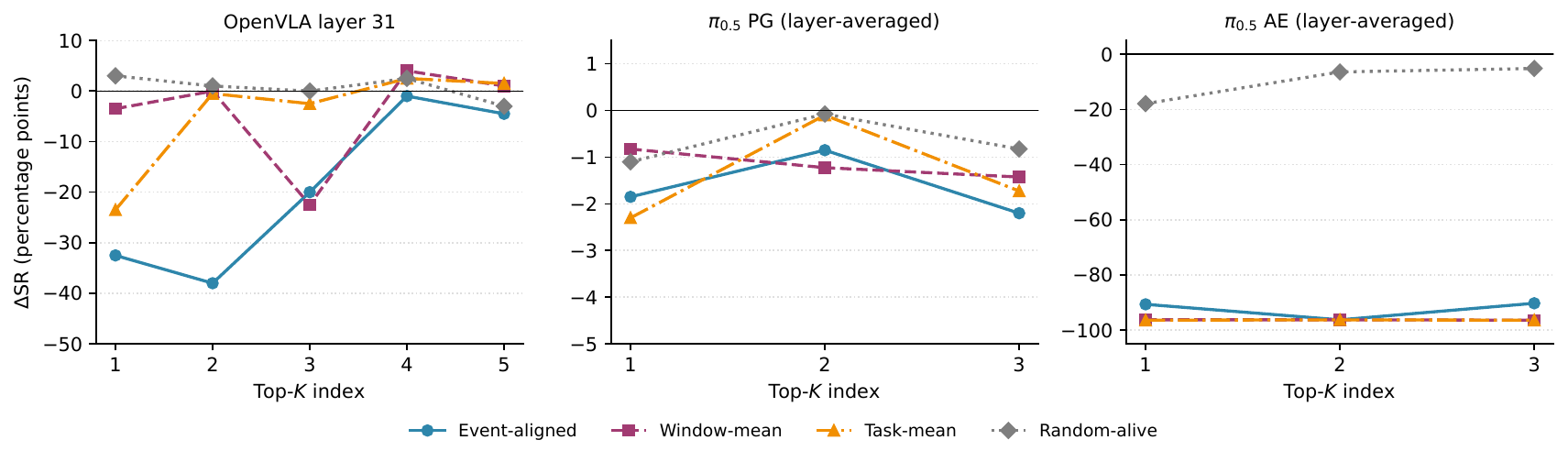}
\caption{
This figure shows per-feature $\Delta\mathrm{SR}$ (percentage points)
under zero-out for the three SAE streams. Left: OpenVLA layer~31
(top-1 to top-5). Middle: $\pi_{0.5}$ PG, layer-averaged across
$\{0, 5, 11, 16\}$ (top-1 to top-3). Right: $\pi_{0.5}$ AE,
layer-averaged across $\{0, 5, 11, 17\}$ (top-1 to top-3). Each
panel uses an independent y-scale to preserve within-stream
structure.
}
\label{fig:topk-three-arch}
\end{figure}

This appendix shows per-feature $\Delta\mathrm{SR}$ across top-$K$
ranks for all three SAE streams. Figure~\ref{fig:topk-three-arch}
visualizes the OpenVLA layer-31 and $\pi_{0.5}$ PG/AE breakdowns. Table~\ref{tab:pi05-topk} reports the corresponding $\pi_{0.5}$ PG and AE
per-feature drops in detail. All values are mean
$\Delta\mathrm{SR}=\mathrm{SR}_{\alpha=0}-\mathrm{SR}_{\mathrm{baseline}}$
in percentage points, averaged over the four LIBERO suites. Negative
values indicate lower success after intervention. Each rank is a separate run that zeroes
only the feature at that rank.

\begin{table}[!ht]
\centering
\small
\begin{subtable}[t]{0.49\linewidth}
\centering
\setlength{\tabcolsep}{3pt}
\begin{tabular}{llrrr}
\toprule
Layer & Ranking & Rank 1 & Rank 2 & Rank 3 \\
\midrule
0  & Event-aligned & $-1.4$ & $-1.4$ & $-3.9$ \\
   & Window-mean   & $-1.4$ & $-1.4$ & $-1.4$ \\
   & Task-mean     & $-1.4$ & $+0.1$ & $-1.4$ \\
   & Random-alive  & $-0.4$ & $+0.6$ & $-2.4$ \\
\cmidrule(lr){1-5}
5  & Event-aligned & $-2.4$ & $+0.1$ & $-0.4$ \\
   & Window-mean   & $-0.4$ & $-0.9$ & $-1.9$ \\
   & Task-mean     & $+0.2$ & $-0.4$ & $-0.4$ \\
   & Random-alive  & $-0.9$ & $+0.2$ & $-0.9$ \\
\cmidrule(lr){1-5}
11 & Event-aligned & $-3.4$ & $-1.4$ & $-3.4$ \\
   & Window-mean   & $-1.9$ & $-1.9$ & $-2.8$ \\
   & Task-mean     & $-4.9$ & $+0.6$ & $-4.4$ \\
   & Random-alive  & $-1.9$ & $-0.9$ & $-0.4$ \\
\cmidrule(lr){1-5}
16 & Event-aligned & $-0.2$ & $-0.7$ & $-1.1$ \\
   & Window-mean   & $+0.4$ & $-0.7$ & $+0.4$ \\
   & Task-mean     & $-3.1$ & $-0.7$ & $-0.7$ \\
   & Random-alive  & $-1.2$ & $-0.2$ & $+0.4$ \\
\bottomrule
\end{tabular}
\subcaption{$\pi_{0.5}$ PG (PaliGemma backbone).}
\label{tab:pi05-topk-pg}
\end{subtable}\hfill
\begin{subtable}[t]{0.49\linewidth}
\centering
\setlength{\tabcolsep}{3pt}
\begin{tabular}{llrrr}
\toprule
Layer & Ranking & Rank 1 & Rank 2 & Rank 3 \\
\midrule
0  & Event-aligned & $-96.4$ & $-96.4$ & $-96.4$ \\
   & Window-mean   & $-96.4$ & $-96.4$ & $-96.4$ \\
   & Task-mean     & $-96.4$ & $-96.4$ & $-96.4$ \\
   & Random-alive  & $+0.6$  & $-1.9$  & $-0.9$  \\
\cmidrule(lr){1-5}
5  & Event-aligned & $-94.4$ & $-95.9$ & $-96.4$ \\
   & Window-mean   & $-96.4$ & $-95.9$ & $-96.4$ \\
   & Task-mean     & $-96.4$ & $-95.9$ & $-96.4$ \\
   & Random-alive  & $-23.9$ & $-24.9$ & $-21.4$ \\
\cmidrule(lr){1-5}
11 & Event-aligned & $-96.4$ & $-96.4$ & $-96.4$ \\
   & Window-mean   & $-95.9$ & $-96.4$ & $-96.4$ \\
   & Task-mean     & $-96.4$ & $-96.4$ & $-96.4$ \\
   & Random-alive  & $-24.9$ & $-0.4$  & $+1.1$  \\
\cmidrule(lr){1-5}
17 & Event-aligned & $-75.4$ & $-96.4$ & $-71.9$ \\
   & Window-mean   & $-96.4$ & $-96.4$ & $-96.4$ \\
   & Task-mean     & $-96.4$ & $-96.4$ & $-96.4$ \\
   & Random-alive  & $-23.4$ & $+1.6$  & $+0.6$  \\
\bottomrule
\end{tabular}
\subcaption{$\pi_{0.5}$ AE (action expert).}
\label{tab:pi05-topk-ae}
\end{subtable}
\caption{Per-feature $\Delta\mathrm{SR}=\mathrm{SR}_{\alpha=0}-\mathrm{SR}_{\mathrm{baseline}}$
(percentage points) for $\pi_{0.5}$ zero-out interventions, broken
down across the top-3 features per ranking and averaged over the
four LIBERO suites. (a) PG drops stay within roughly $5$ pp across
all rows. (b) AE event-aligned, window-mean, and task-mean entries
cluster near the policy-collapse floor ($\approx-96$), leaving the
differences among the top-3 features largely indistinguishable.}
\label{tab:pi05-topk}
\end{table}
\FloatBarrier

\section{Effects of Feature Suppression on Target and Off-Target Tasks}
\label{app:target-offtarget-zeroout}

We corrected a task mapping error
in the original OpenVLA evaluation and extended the target/off-target
comparison to $\pi_{0.5}$. To test whether event-selected features have
task-specific effects, we select the highest-scoring feature from each of
30 sampled LIBERO-10 event clusters per model. We suppress each feature on
the task that supplied the cluster (target) and on a different task
(off-target), using 15 initial states per task. We zero out OpenVLA layer-31
features and scale $\pi_{0.5}$ action-expert layer-11 features by
$\alpha=0.7$ to avoid saturation, following the strength sweep in
Figure~\ref{fig:ae-libero10-alpha-sweep}. For each task, we compare success
with unmodified rollouts from the same initial states.

\begin{table}[!ht]
\centering
\small
\begin{tabular}{lccc}
\toprule
Model (site, $\alpha$) & Target SR & Off-target SR & Difference (pp) \\
\midrule
OpenVLA L31, $\alpha=0$ & 37.6\% (55.3\%) & 40.2\% (57.6\%) & $-0.4$ [$-6.9$, $6.0$] \\
$\pi_{0.5}$ AE L11, $\alpha=0.7$ & 70.7\% (94.7\%) & 70.7\% (97.6\%) & $+2.9$ [$-1.6$, $7.3$] \\
\bottomrule
\end{tabular}
\caption{This table compares target and off-target suppression on LIBERO-10.
Parentheses give baseline SRs from the same initial states. Difference is
target minus off-target SR change, with 95\% bootstrap confidence intervals, computed by resampling within each task, in brackets.}
\label{tab:target-offtarget-zeroout}
\end{table}

The
updated results support the same qualitative conclusion as originally
reported. Target and off-target SR drops are similar: 17.8 versus 17.3
percentage points for OpenVLA and 24.0 versus 26.9 for the $\pi_{0.5}$ action
expert (Table~\ref{tab:target-offtarget-zeroout}). This pattern suggests that many
event-selected features participate in manipulation roles shared across
tasks, such as reaching, grasping, transport, and release.

\FloatBarrier
\section{Comparison with Non-SAE Interventions}
\label{app:steering-safety-diagnostics}

We test whether directions obtained without an SAE have similar behavioral
effects when the perturbation magnitude matches SAE feature suppression. We
evaluate OpenVLA residual layer 31 and the $\pi_{0.5}$ action expert at layer 17
on five tasks from each of the four LIBERO suites, with ten initial states per
task. The SAEs for this comparison are trained separately, on the same hidden
states used to fit PCA and ICA.

We compare six interventions with the unmodified policy (Table~\ref{tab:nonsae-controls}). Each direction is
selected per task before evaluation, from activation traces averaged within
each environment step.
\begin{description}[leftmargin=0pt,itemsep=2pt,parsep=0pt]
\item[Selected SAE feature.] We select the highest-scoring feature
in the full dictionary using event-aligned ranking and scale its activation
by $\alpha=0.6$. The residual-preserving edit in
Section~\ref{sec:intervention} gives the modified hidden state
$\widetilde{\mathbf{h}}$.
\item[Hidden coordinate.] We rank the activation traces of all
hidden coordinates and select one coordinate axis.
\item[PCA direction.] We fit the 64 principal components with
largest variance in centered hidden states and rank their projection traces.
\item[ICA direction.] We whiten the 64 PCA components and fit
64 components with FastICA. We rank their activation traces and use the
selected component's mixing direction in the original hidden space.
\item[Random orthogonal direction.] We use QR decomposition of
64 Gaussian vectors to construct 64 mutually orthogonal unit directions
and rank their projection traces.
\item[Difference in means.] We subtract the mean hidden state
outside event windows from the mean inside event windows, giving one
direction per task without ranking.
\end{description}

For each non-SAE direction, let $\mathbf{v}$ be its selected unit direction, with
its sign chosen so that projections are higher on average inside event windows
than outside. At the current hidden state $\mathbf{h}$, we compute the magnitude
of the counterfactual SAE edit, $\delta=\|\widetilde{\mathbf{h}}-\mathbf{h}\|_2$,
and apply the update in Equation~\ref{eq:matched-norm-control}, recomputing
$\delta$ at every step.
\begin{equation}
\mathbf{h}'=\mathbf{h}-\delta\mathbf{v}.
\label{eq:matched-norm-control}
\end{equation}

\begin{table}[!ht]
\centering\small
\setlength{\tabcolsep}{5pt}
\begin{tabular}{lcc}
\toprule
Intervention & OpenVLA L31 SR (\%) & $\pi_{0.5}$ AE L17 SR (\%) \\
\midrule
No edit & 72.0 [66.5, 77.5] & 99.0 [97.5, 100.0] \\
Selected SAE feature & 72.5 [67.0, 78.0] & 77.0 [74.0, 80.0] \\
Hidden coordinate & 71.0 [65.5, 76.5] & 83.0 [81.0, 84.5] \\
PCA direction & 73.0 [67.5, 78.5] & 79.5 [78.5, 80.0] \\
ICA direction & 73.0 [67.5, 78.5] & 84.0 [82.5, 85.0] \\
Random orthogonal & 70.5 [65.0, 76.0] & 89.0 [86.0, 91.5] \\
Difference in means & 72.5 [67.0, 78.0] & 90.5 [87.5, 93.0] \\
\bottomrule
\end{tabular}
\caption{Success rates when SAE and non-SAE directions are edited with
matched perturbation norms, with 95\% bootstrap confidence intervals in brackets. Each model is evaluated on 20 tasks
with ten initial states per task, and SAE suppression uses $\alpha=0.6$.}
\label{tab:nonsae-controls}
\end{table}

In the $\pi_{0.5}$ action expert, suppressing the selected SAE feature reduces
success the most, and every non-SAE direction also reduces success. In OpenVLA,
no intervention moves success outside the confidence interval of the unmodified
policy.

\FloatBarrier
\section{Ranking and SAE Stability}
\label{app:ranking-stability}

\textbf{Sampling and clustering stability.} We test whether changing the
rollout sample changes which features rank highly on LIBERO-Spatial, keeping
each SAE fixed. With event clusters fixed, we draw rollout samples with
replacement within each task and recompute the rankings 1,000 times. In a
second test, we randomly retain 80\% of the rollouts from each task and repeat
both clustering and ranking, doing this 200 times. We compare each resulting
ranking with the ranking computed from all original rollouts.

Rankings change little in both tests, with PaliGemma rankings changing
the least (Table~\ref{tab:sampling-ranking-stability}). For the action expert,
however, similar overall rankings do not always select the same top features:
the new top-3 typically includes 2 of the original top-3 features.

\begin{table}[!htbp]
\centering\small
\setlength{\tabcolsep}{5pt}
\begin{tabular}{lccccc}
\toprule
& & \multicolumn{2}{c}{Fixed clusters} & \multicolumn{2}{c}{Reclustering} \\
\cmidrule(lr){3-4}\cmidrule(lr){5-6}
Model/component & $K$ & Spearman & Top-$K$ retained & Spearman & Top-$K$ retained \\
\midrule
OpenVLA layer 31 & 5 & 0.989 & 5 [4, 5] & 0.994 & 5 [4, 5] \\
$\pi_{0.5}$ PaliGemma layer 5 & 3 & 0.997 & 3 [3, 3] & 0.999 & 3 [3, 3] \\
$\pi_{0.5}$ action expert layer 17 & 3 & 0.951 & 2 [2, 3] & 0.977 & 2 [2, 3] \\
\bottomrule
\end{tabular}
\caption{Event-aligned ranking stability on LIBERO-Spatial. The table reports
median values over repeated sampling. Top-$K$ retained counts how many original
top-$K$ features are selected again, with brackets showing the middle 95\% of
results.}
\label{tab:sampling-ranking-stability}
\end{table}

\textbf{SAE training stability.} We train 3 SAEs with different random
seeds on LIBERO-Goal at each of two sites: OpenVLA layer 31 and the
$\pi_{0.5}$ action-expert layer 17. For OpenVLA, FVE ranges from 0.9950 to
0.9951, mean $L_0$ from 64.48 to 64.75, and the alive-feature fraction from
9.6\% to 10.3\%. For the action expert, the corresponding ranges are
0.9602--0.9608, 63.96--64.00, and 94.4\%--95.5\%. We also evaluate one SAE
per site with a reconstruction-only hook. OpenVLA achieves a Hooked SR of
38\%, compared with a Raw SR of 65\%. The action expert achieves 95\%,
compared with 97\%. To further assess variation across training seeds, we additionally
evaluate separately trained SAEs and the unmodified policy on the same 100
LIBERO-Goal task and initial-state assignments.
Hooked SR is $32 \pm 2.6\%$ for OpenVLA and $98 \pm 1.0\%$ for the action
expert (mean $\pm$ standard deviation across training seeds). The corresponding
Raw SRs are 76\% and 98\%, respectively. Offline diagnostics and Hooked SR
therefore remain stable across training seeds.

We assess ranking stability across SAE training seeds. Sparse
decompositions are not unique: one SAE may represent a direction with a single
feature, while another represents it with several correlated
features~\citep{bricken2023towards,elhage2022toy,chanin2025a}. We therefore
match features across SAEs by their decoder directions and compare their
event-aligned rankings. Across seed pairs, 0--2 of the top-5 features match.
Across the 10 LIBERO-Goal tasks, the median pairwise Spearman correlation is
0.917 (range 0.903--0.923) for OpenVLA and 0.891 (range 0.873--0.920) for the
action expert. We further test whether the action-expert suppression in
Section~\ref{sec:exp-causal} persists on new initial states and with
independently trained SAEs. The original LIBERO-Goal experiment evaluated
five initial states per task. Here we use ten new initial states per task,
giving 100 task and initial-state assignments that are also disjoint from
those used for ranking. We use the original $\pi_{0.5}$ action-expert
layer-17 SAE and two additional SAEs trained with different random seeds.
For each SAE, we independently select its top-3 features and evaluate
each feature's zero-out alongside the unmodified policy on these assignments.
For the original SAE, selection recovers the same top-3 features, and the two
features that previously caused complete failure again give 0\% SR. Across
the three SAEs, mean SR reductions range from 67.0 to 88.7 percentage
points, compared with 69.4 points on LIBERO-Goal in the original
experiment. The action-expert suppression effect therefore persists on new
initial states and across independently trained SAEs.

\FloatBarrier
\section{Finite-Window Feature Suppression}
\label{app:windowed-suppression}

The experiments in Section~\ref{sec:exp-causal} apply feature edits
throughout each rollout. Persistent perturbations may accumulate and drive the
policy toward states rarely visited during unmodified rollouts. We compare
full-rollout suppression with suppression in shorter windows to examine how
intervention timing and duration affect task success.

We test the $\pi_{0.5}$ action expert at layer 17 on three tasks from each
of four LIBERO suites. Each task uses 15 initial states where the unmodified
policy succeeds and one top-ranked feature selected by event-aligned ranking
(Section~\ref{sec:feature-ranking}). All interventions use $\alpha=0$. Windows
are fixed before intervention using the corresponding successful unmodified
rollout as a reference. Table~\ref{tab:windowed-suppression} compares five
conditions.
\begin{description}[leftmargin=0pt,itemsep=2pt,parsep=0pt]
\item[Before completion (10 steps).] We suppress the selected feature
in a window ending at reference task completion, with a planned duration of
10 environment steps.
\item[Before completion (30 steps).] We extend this window backward
to 30 steps, keeping its endpoint and the selected feature unchanged.
\item[Earlier (30 steps).] We choose a 30-step interval entirely
before the final 30-step window, maximizing summed activation of the selected
feature in the reference rollout. We suppress the feature within this earlier
interval.
\item[Full rollout.] We suppress the same selected feature
throughout the rollout.
\item[Random (30 steps).] We suppress one fixed random feature per
suite in the same final 30-step window used for the selected feature.
\end{description}

\begin{table}[!htbp]
\centering
\small
\setlength{\tabcolsep}{4pt}
\begin{tabular}{lcrrrrr}
\toprule
Suite & Task ID & \multicolumn{2}{c}{Before completion} & Earlier & Full rollout & Random \\
\cmidrule(lr){3-4}
 & & 10 steps & 30 steps & 30 steps & & 30 steps \\
\midrule
Spatial & 1 & 53.3\% & 46.7\% & 0.0\% & 0.0\% & 100.0\% \\
 & 5 & 86.7\% & 73.3\% & 33.3\% & 0.0\% & 100.0\% \\
 & 7 & 100.0\% & 100.0\% & 100.0\% & 100.0\% & 100.0\% \\
\addlinespace[2pt]
Object & 1 & 100.0\% & 93.3\% & 73.3\% & 0.0\% & 100.0\% \\
 & 5 & 86.7\% & 60.0\% & 6.7\% & 0.0\% & 86.7\% \\
 & 7 & 100.0\% & 100.0\% & 86.7\% & 0.0\% & 100.0\% \\
\addlinespace[2pt]
Goal & 1 & 100.0\% & 100.0\% & 100.0\% & 100.0\% & 100.0\% \\
 & 5 & 80.0\% & 40.0\% & 80.0\% & 0.0\% & 100.0\% \\
 & 7 & 93.3\% & 93.3\% & 26.7\% & 0.0\% & 100.0\% \\
\addlinespace[2pt]
LIBERO-10 & 1 & 66.7\% & 33.3\% & 80.0\% & 0.0\% & 93.3\% \\
 & 5 & 40.0\% & 13.3\% & 13.3\% & 0.0\% & 26.7\% \\
 & 7 & 100.0\% & 100.0\% & 100.0\% & 0.0\% & 93.3\% \\
\midrule
\multicolumn{2}{l}{Mean} & 83.9\% & 71.1\% & 58.3\% & 16.7\% & 91.7\% \\
\bottomrule
\end{tabular}
\caption{Task success under full and windowed feature suppression
($\alpha=0$) in the $\pi_{0.5}$ action expert at layer 17. Each entry uses 15
initial states on which the unmodified policy succeeds. Window durations denote
planned environment steps. Task IDs are zero-based within each suite.}
\label{tab:windowed-suppression}
\end{table}

From Table~\ref{tab:windowed-suppression}, we observe that intervening at
different stages of a rollout can produce different effects on task success.
Across the tested tasks, full-rollout suppression is more disruptive on
average than finite-window suppression. These observations motivate future
work on adaptive steering that adjusts intervention timing and duration to the
policy's ongoing behavior.

\FloatBarrier
\section{Effects of Feature Edits on Robot Motion}
\label{app:positive-steering}

We report additional experiments conducted during the rebuttal phase
on object lift and end-effector motion. The lift experiments use different SAEs
and features from those in Figure~\ref{fig:bowl-lift-search}. We vary the activation scale from $0\times$ to $2\times$
for two features selected from event-aligned rankings at $\pi_{0.5}$ action-expert layer 5: feature 21
on LIBERO-Goal ``put the bowl on the stove'' and feature 628 on LIBERO-Object
``pick up the milk and place it in the basket.'' At each scale, we measure the
maximum height each object is lifted on the same 30 previously unused initial
states per task.

For feature 21, bowl lift shows an overall upward trend as the scale
increases through $1.5\times$ (Holm-adjusted
$p=2\times10^{-5}$), but does not increase at every scale
(Table~\ref{tab:bowl-dose-response}). At $1.5\times$, maximum bowl lift is
0.945~cm higher than with no intervention (95\% CI $[0.431, 1.458]$) and
2.244~cm higher than when a random feature is amplified by the same factor
($[1.621, 2.825]$), with no reliable changes in task success, timeout, or
trajectory instability. At the same scale, amplifying feature 628 increases
milk lift over no intervention but not over amplification of a random feature
($-0.398$~cm, $[-0.825, 0.021]$). Together with the main-text experiments,
these results show that feature amplification can adjust robot motion, while
reliably selecting a feature and scale for a desired change remains an open
problem.

\begin{table}[!htbp]
\centering
\small
\begin{tabular}{lcc}
\toprule
Scale & Bowl lift vs.\ $1\times$ (cm, 95\% CI) & SR \\
\midrule
$0\times$ & $-14.080$ $[-14.464, -13.676]$ & 0/30 \\
$0.5\times$ & $-0.461$ $[-0.944, 0.035]$ & 29/30 \\
$1\times$ (unchanged) & 0 & 30/30 \\
$1.5\times$ & $+0.945$ $[0.431, 1.458]$ & 29/30 \\
$2\times$ & $+0.594$ $[0.092, 1.106]$ & 30/30 \\
\bottomrule
\end{tabular}
\caption{Maximum bowl lift under scaling of $\pi_{0.5}$ action-expert
layer-5 feature 21 on LIBERO-Goal ``put the bowl on the stove,'' relative to
the same initial state with no intervention. Each scale uses 30
previously unused initial states.}
\label{tab:bowl-dose-response}
\end{table}

\textbf{End-effector motion.} In OpenVLA layer 31 on LIBERO-Spatial, we measure
how far the end-effector trajectory deviates from the unmodified rollout on the
same task and initial state. Across 250 feature--episode pairs per group,
zeroing event-aligned features causes 5.37~cm more deviation than
zeroing random-alive features (95\% CI $[4.69, 6.09]$). Among rollout pairs where
both the unmodified and edited rollouts succeed, mean deviation is 4.47~cm for
event-aligned features and 1.84~cm for random-alive features, computed over each
group's successful pairs.
Feature edits can therefore change the trajectory even when the task succeeds.
Our zero-out results show that changing a feature
can change policy behavior, but do not establish that the feature alone produces
its assigned behavior.

\textbf{Amplification under an unrelated instruction.} The lift experiments
above adjust a motion the policy already performs. To test whether amplification
can produce the motion without the task instruction, we use an unrelated
non-manipulation instruction. The unmodified policy still completed ``put the
bowl on the stove'' in 9 of 30 LIBERO-Goal rollouts. Amplifying feature 21 did
not increase bowl lift over no intervention ($-4.124$~cm, 95\% CI
$[-7.090, -1.074]$) or over the mean of ten matched random features for each
initial state ($+0.587$~cm, $[-1.094, 2.335]$). We therefore do not claim
instruction-free induction.
\FloatBarrier
\section{Success Prediction and Hook-Site Comparison}
\label{app:sanity-checks}

\paragraph{Success predictability of mean-pooled SAE codes in OpenVLA.}
\label{app:predictive-fidelity}
As a static sanity check on top of the reconstruction results in
Section~\ref{sec:exp-sae-reconstruction}, we ask whether mean-pooled
SAE activations carry rollout-level outcome information. For each
rollout we form $\bar{z} = (1/T)\sum_{t=1}^{T} z_t \in \mathbb{R}^{m}$
and train an $L_2$-regularized logistic-regression probe to predict
the binary success label with 5-fold stratified cross-validation.
Table~\ref{tab:predictive-fidelity} compares the SAE probe against
raw hidden states, task-identity,
and shuffled labels. The picture is
consistent with reconstruction fidelity: SAE codes retain
behaviorally-relevant signal but are not lossless.

\begin{table}[!ht]
\centering
\small
\begin{tabular}{lcccc}
\toprule
Suite & SAE & Raw hidden & Task-id only & Shuffled labels \\
\midrule
LIBERO-Spatial & 0.794 & 0.976 & 0.550 & 0.498 \\
LIBERO-Object  & 0.926 & 0.942 & 0.636 & 0.498 \\
LIBERO-Goal    & 0.837 & 0.896 & 0.627 & 0.576 \\
LIBERO-10      & 0.791 & 0.893 & 0.538 & 0.484 \\
\bottomrule
\end{tabular}
\caption{Balanced accuracy of trajectory-level success-prediction
probes (5-fold cross-validation).}
\label{tab:predictive-fidelity}
\end{table}

\paragraph{Behavioral comparison of intervention hook sites.}
We also compare two natural SAE hook sites on OpenVLA LIBERO-Spatial
layer~31: the MLP output (after the MLP block) and
the residual stream (after the residual addition; used throughout
the main paper). Table~\ref{tab:mlp-out-resid-post-behavior} reports
Hooked SR alongside Zero-out SR.
The MLP-output SAE preserves higher fidelity while
the residual-stream SAE yields a larger SR drop under zero-out. We use the residual stream for the main experiments,
given our focus on behavioral sensitivity to single-feature edits,
consistent with work on residual-stream SAEs in language
models~\citep{templeton2024scaling, gao2025scaling} and
VLAs~\citep{khan2025controlling, swann2026sparse}.

\begin{table}[!ht]
\centering
\small
\begin{tabular}{lccc}
\toprule
Hook site & Layer & Hooked SR $\uparrow$ & Zero-out SR $\downarrow$ \\
\midrule
MLP output    & 31 & $0.83 \pm 0.17$  & $0.752 \pm 0.060$ \\
Residual stream & 31 & $0.67 \pm 0.25$  & $0.516 \pm 0.272$ \\
\bottomrule
\end{tabular}
\caption{Behavioral comparison of SAE intervention sites on
LIBERO-Spatial. Hooked SR is mean $\pm$ SD across tasks (10 rollouts per task). Zero-out SR is mean $\pm$ SD across the top-5 event-aligned features, each zeroed separately ($\alpha=0$) with 5 rollouts per task.}
\label{tab:mlp-out-resid-post-behavior}
\end{table}

\newpage
\input{checklist.tex}

\end{document}

%% file: checklist.tex
\section*{NeurIPS Paper Checklist}

\begin{enumerate}

\item {\bf Claims}
    \item[] Question: Do the main claims made in the abstract and introduction accurately reflect the paper's contributions and scope?
    \item[] Answer: \answerYes{} %
    \item[] Justification: The claims in the abstract and Section~\ref{sec:intro} are empirically supported by the experiments in Section~\ref{sec:experiment}.
    \item[] Guidelines:
    \begin{itemize}
        \item The answer \answerNA{} means that the abstract and introduction do not include the claims made in the paper.
        \item The abstract and/or introduction should clearly state the claims made, including the contributions made in the paper and important assumptions and limitations. A \answerNo{} or \answerNA{} answer to this question will not be perceived well by the reviewers. 
        \item The claims made should match theoretical and experimental results, and reflect how much the results can be expected to generalize to other settings. 
        \item It is fine to include aspirational goals as motivation as long as it is clear that these goals are not attained by the paper. 
    \end{itemize}

\item {\bf Limitations}
    \item[] Question: Does the paper discuss the limitations of the work performed by the authors?
    \item[] Answer: \answerYes{} %
    \item[] Justification: Section~\ref{sec:limitations} discusses limitations in isolating individual behavioral variables, interpreting SAE feature semantics, and evaluating behavior with binary success. The cost of closed-loop rollouts also limits test coverage.
    \item[] Guidelines:
    \begin{itemize}
        \item The answer \answerNA{} means that the paper has no limitation while the answer \answerNo{} means that the paper has limitations, but those are not discussed in the paper. 
        \item The authors are encouraged to create a separate ``Limitations'' section in their paper.
        \item The paper should point out any strong assumptions and how robust the results are to violations of these assumptions (e.g., independence assumptions, noiseless settings, model well-specification, asymptotic approximations only holding locally). The authors should reflect on how these assumptions might be violated in practice and what the implications would be.
        \item The authors should reflect on the scope of the claims made, e.g., if the approach was only tested on a few datasets or with a few runs. In general, empirical results often depend on implicit assumptions, which should be articulated.
        \item The authors should reflect on the factors that influence the performance of the approach. For example, a facial recognition algorithm may perform poorly when image resolution is low or images are taken in low lighting. Or a speech-to-text system might not be used reliably to provide closed captions for online lectures because it fails to handle technical jargon.
        \item The authors should discuss the computational efficiency of the proposed algorithms and how they scale with dataset size.
        \item If applicable, the authors should discuss possible limitations of their approach to address problems of privacy and fairness.
        \item While the authors might fear that complete honesty about limitations might be used by reviewers as grounds for rejection, a worse outcome might be that reviewers discover limitations that aren't acknowledged in the paper. The authors should use their best judgment and recognize that individual actions in favor of transparency play an important role in developing norms that preserve the integrity of the community. Reviewers will be specifically instructed to not penalize honesty concerning limitations.
    \end{itemize}

\item {\bf Theory assumptions and proofs}
    \item[] Question: For each theoretical result, does the paper provide the full set of assumptions and a complete (and correct) proof?
    \item[] Answer: \answerNA{} %
    \item[] Justification: The paper presents an empirical pipeline and intervention study and contains no theoretical results requiring assumptions or proofs.
    \item[] Guidelines:
    \begin{itemize}
        \item The answer \answerNA{} means that the paper does not include theoretical results. 
        \item All the theorems, formulas, and proofs in the paper should be numbered and cross-referenced.
        \item All assumptions should be clearly stated or referenced in the statement of any theorems.
        \item The proofs can either appear in the main paper or the supplemental material, but if they appear in the supplemental material, the authors are encouraged to provide a short proof sketch to provide intuition. 
        \item Inversely, any informal proof provided in the core of the paper should be complemented by formal proofs provided in appendix or supplemental material.
        \item Theorems and Lemmas that the proof relies upon should be properly referenced. 
    \end{itemize}

    \item {\bf Experimental result reproducibility}
    \item[] Question: Does the paper fully disclose all the information needed to reproduce the main experimental results of the paper to the extent that it affects the main claims and/or conclusions of the paper (regardless of whether the code and data are provided or not)?
    \item[] Answer: \answerYes{} %
    \item[] Justification: The pipeline is described in Section~\ref{sec:method}, with hyperparameters and per-stream details in the appendices. We use openly available pretrained policies (OpenVLA, $\pi_{0.5}$) and the open LIBERO benchmark.
    \item[] Guidelines:
    \begin{itemize}
        \item The answer \answerNA{} means that the paper does not include experiments.
        \item If the paper includes experiments, a \answerNo{} answer to this question will not be perceived well by the reviewers: Making the paper reproducible is important, regardless of whether the code and data are provided or not.
        \item If the contribution is a dataset and\slash or model, the authors should describe the steps taken to make their results reproducible or verifiable. 
        \item Depending on the contribution, reproducibility can be accomplished in various ways. For example, if the contribution is a novel architecture, describing the architecture fully might suffice, or if the contribution is a specific model and empirical evaluation, it may be necessary to either make it possible for others to replicate the model with the same dataset, or provide access to the model. In general. releasing code and data is often one good way to accomplish this, but reproducibility can also be provided via detailed instructions for how to replicate the results, access to a hosted model (e.g., in the case of a large language model), releasing of a model checkpoint, or other means that are appropriate to the research performed.
        \item While NeurIPS does not require releasing code, the conference does require all submissions to provide some reasonable avenue for reproducibility, which may depend on the nature of the contribution. For example
        \begin{enumerate}
            \item If the contribution is primarily a new algorithm, the paper should make it clear how to reproduce that algorithm.
            \item If the contribution is primarily a new model architecture, the paper should describe the architecture clearly and fully.
            \item If the contribution is a new model (e.g., a large language model), then there should either be a way to access this model for reproducing the results or a way to reproduce the model (e.g., with an open-source dataset or instructions for how to construct the dataset).
            \item We recognize that reproducibility may be tricky in some cases, in which case authors are welcome to describe the particular way they provide for reproducibility. In the case of closed-source models, it may be that access to the model is limited in some way (e.g., to registered users), but it should be possible for other researchers to have some path to reproducing or verifying the results.
        \end{enumerate}
    \end{itemize}

\item {\bf Open access to data and code}
    \item[] Question: Does the paper provide open access to the data and code, with sufficient instructions to faithfully reproduce the main experimental results, as described in supplemental material?
    \item[] Answer: \answerYes{} %
    \item[] Justification: Code, intervention scripts, and links to trained SAE checkpoints are publicly available at \url{https://github.com/xc-j/Event-SAE}. The underlying pretrained policies (OpenVLA, $\pi_{0.5}$) and the LIBERO benchmark are already openly available.
    \item[] Guidelines:
    \begin{itemize}
        \item The answer \answerNA{} means that paper does not include experiments requiring code.
        \item Please see the NeurIPS code and data submission guidelines (\url{https://neurips.cc/public/guides/CodeSubmissionPolicy}) for more details.
        \item While we encourage the release of code and data, we understand that this might not be possible, so \answerNo{} is an acceptable answer. Papers cannot be rejected simply for not including code, unless this is central to the contribution (e.g., for a new open-source benchmark).
        \item The instructions should contain the exact command and environment needed to run to reproduce the results. See the NeurIPS code and data submission guidelines (\url{https://neurips.cc/public/guides/CodeSubmissionPolicy}) for more details.
        \item The authors should provide instructions on data access and preparation, including how to access the raw data, preprocessed data, intermediate data, and generated data, etc.
        \item The authors should provide scripts to reproduce all experimental results for the new proposed method and baselines. If only a subset of experiments are reproducible, they should state which ones are omitted from the script and why.
        \item At submission time, to preserve anonymity, the authors should release anonymized versions (if applicable).
        \item Providing as much information as possible in supplemental material (appended to the paper) is recommended, but including URLs to data and code is permitted.
    \end{itemize}

\item {\bf Experimental setting/details}
    \item[] Question: Does the paper specify all the training and test details (e.g., data splits, hyperparameters, how they were chosen, type of optimizer) necessary to understand the results?
    \item[] Answer: \answerYes{} %
    \item[] Justification: Setup is described in Section~\ref{sec:exp:setup}, with full hyperparameters and configurations in the appendices (SAE training, clustering, VLM prompt, real-robot setup).
    \item[] Guidelines:
    \begin{itemize}
        \item The answer \answerNA{} means that the paper does not include experiments.
        \item The experimental setting should be presented in the core of the paper to a level of detail that is necessary to appreciate the results and make sense of them.
        \item The full details can be provided either with the code, in appendix, or as supplemental material.
    \end{itemize}

\item {\bf Experiment statistical significance}
    \item[] Question: Does the paper report error bars suitably and correctly defined or other appropriate information about the statistical significance of the experiments?
    \item[] Answer: \answerYes{} %
    \item[] Justification: Standard deviations are reported in Figure~\ref{fig:real-robot-steering-strength}, Tables~\ref{tab:setup}, \ref{tab:pg-active-drop16}, and~\ref{tab:mlp-out-resid-post-behavior}, and Appendix~\ref{app:ranking-stability}. Confidence intervals are reported in Figure~\ref{fig:feature-scaling}, Tables~\ref{tab:target-offtarget-zeroout} and~\ref{tab:nonsae-controls}, and Appendix~\ref{app:positive-steering}. Table~\ref{tab:sampling-ranking-stability} reports central 95\% ranges of top-$K$ retention over repeated sampling.
    \item[] Guidelines:
    \begin{itemize}
        \item The answer \answerNA{} means that the paper does not include experiments.
        \item The authors should answer \answerYes{} if the results are accompanied by error bars, confidence intervals, or statistical significance tests, at least for the experiments that support the main claims of the paper.
        \item The factors of variability that the error bars are capturing should be clearly stated (for example, train/test split, initialization, random drawing of some parameter, or overall run with given experimental conditions).
        \item The method for calculating the error bars should be explained (closed form formula, call to a library function, bootstrap, etc.)
        \item The assumptions made should be given (e.g., Normally distributed errors).
        \item It should be clear whether the error bar is the standard deviation or the standard error of the mean.
        \item It is OK to report 1-sigma error bars, but one should state it. The authors should preferably report a 2-sigma error bar than state that they have a 96\% CI, if the hypothesis of Normality of errors is not verified.
        \item For asymmetric distributions, the authors should be careful not to show in tables or figures symmetric error bars that would yield results that are out of range (e.g., negative error rates).
        \item If error bars are reported in tables or plots, the authors should explain in the text how they were calculated and reference the corresponding figures or tables in the text.
    \end{itemize}

\item {\bf Experiments compute resources}
    \item[] Question: For each experiment, does the paper provide sufficient information on the computer resources (type of compute workers, memory, time of execution) needed to reproduce the experiments?
    \item[] Answer: \answerYes{} %
    \item[] Justification: Appendix~\ref{app:sae-training} reports NVIDIA A100 GPUs, a few GPU-hours of training per SAE, and rollout budgets for the main experiments.
    \item[] Guidelines:
    \begin{itemize}
        \item The answer \answerNA{} means that the paper does not include experiments.
        \item The paper should indicate the type of compute workers CPU or GPU, internal cluster, or cloud provider, including relevant memory and storage.
        \item The paper should provide the amount of compute required for each of the individual experimental runs as well as estimate the total compute. 
        \item The paper should disclose whether the full research project required more compute than the experiments reported in the paper (e.g., preliminary or failed experiments that didn't make it into the paper). 
    \end{itemize}
    
\item {\bf Code of ethics}
    \item[] Question: Does the research conducted in the paper conform, in every respect, with the NeurIPS Code of Ethics \url{https://neurips.cc/public/EthicsGuidelines}?
    \item[] Answer: \answerYes{} %
    \item[] Justification: The research conforms with the NeurIPS Code of Ethics. It uses publicly available models and benchmarks. The real-robot study runs in a controlled lab setting with no human subjects or personally identifying data.
    \item[] Guidelines:
    \begin{itemize}
        \item The answer \answerNA{} means that the authors have not reviewed the NeurIPS Code of Ethics.
        \item If the authors answer \answerNo, they should explain the special circumstances that require a deviation from the Code of Ethics.
        \item The authors should make sure to preserve anonymity (e.g., if there is a special consideration due to laws or regulations in their jurisdiction).
    \end{itemize}

\item {\bf Broader impacts}
    \item[] Question: Does the paper discuss both potential positive societal impacts and negative societal impacts of the work performed?
    \item[] Answer: \answerYes{} %
    \item[] Justification: Our work is among the early studies of mechanistic interpretability in VLA policies. Positive impacts include better understanding of robot policies and the potential to adjust robot behavior through feature-level interventions, which may support more transparent control and safer deployment. Potential negative impacts include safety risks from poorly tuned interventions, which can produce large robot motions (Section~\ref{sec:exp-real-robot}).
    \item[] Guidelines:
    \begin{itemize}
        \item The answer \answerNA{} means that there is no societal impact of the work performed.
        \item If the authors answer \answerNA{} or \answerNo, they should explain why their work has no societal impact or why the paper does not address societal impact.
        \item Examples of negative societal impacts include potential malicious or unintended uses (e.g., disinformation, generating fake profiles, surveillance), fairness considerations (e.g., deployment of technologies that could make decisions that unfairly impact specific groups), privacy considerations, and security considerations.
        \item The conference expects that many papers will be foundational research and not tied to particular applications, let alone deployments. However, if there is a direct path to any negative applications, the authors should point it out. For example, it is legitimate to point out that an improvement in the quality of generative models could be used to generate Deepfakes for disinformation. On the other hand, it is not needed to point out that a generic algorithm for optimizing neural networks could enable people to train models that generate Deepfakes faster.
        \item The authors should consider possible harms that could arise when the technology is being used as intended and functioning correctly, harms that could arise when the technology is being used as intended but gives incorrect results, and harms following from (intentional or unintentional) misuse of the technology.
        \item If there are negative societal impacts, the authors could also discuss possible mitigation strategies (e.g., gated release of models, providing defenses in addition to attacks, mechanisms for monitoring misuse, mechanisms to monitor how a system learns from feedback over time, improving the efficiency and accessibility of ML).
    \end{itemize}
    
\item {\bf Safeguards}
    \item[] Question: Does the paper describe safeguards that have been put in place for responsible release of data or models that have a high risk for misuse (e.g., pre-trained language models, image generators, or scraped datasets)?
    \item[] Answer: \answerNA{} %
    \item[] Justification: The paper does not release high-risk generative models or scraped datasets. Released artifacts are interpretability tooling for existing public VLA policies (OpenVLA, $\pi_{0.5}$).
    \item[] Guidelines:
    \begin{itemize}
        \item The answer \answerNA{} means that the paper poses no such risks.
        \item Released models that have a high risk for misuse or dual-use should be released with necessary safeguards to allow for controlled use of the model, for example by requiring that users adhere to usage guidelines or restrictions to access the model or implementing safety filters. 
        \item Datasets that have been scraped from the Internet could pose safety risks. The authors should describe how they avoided releasing unsafe images.
        \item We recognize that providing effective safeguards is challenging, and many papers do not require this, but we encourage authors to take this into account and make a best faith effort.
    \end{itemize}

\item {\bf Licenses for existing assets}
    \item[] Question: Are the creators or original owners of assets (e.g., code, data, models), used in the paper, properly credited and are the license and terms of use explicitly mentioned and properly respected?
    \item[] Answer: \answerYes{} %
    \item[] Justification: All external assets (pretrained policies OpenVLA and $\pi_{0.5}$, the SigLIP vision encoder, the LIBERO benchmark, the AWE algorithm, and the Mobile ALOHA hardware platform) are cited at first use and used in accordance with their original licenses.
    \item[] Guidelines:
    \begin{itemize}
        \item The answer \answerNA{} means that the paper does not use existing assets.
        \item The authors should cite the original paper that produced the code package or dataset.
        \item The authors should state which version of the asset is used and, if possible, include a URL.
        \item The name of the license (e.g., CC-BY 4.0) should be included for each asset.
        \item For scraped data from a particular source (e.g., website), the copyright and terms of service of that source should be provided.
        \item If assets are released, the license, copyright information, and terms of use in the package should be provided. For popular datasets, \url{paperswithcode.com/datasets} has curated licenses for some datasets. Their licensing guide can help determine the license of a dataset.
        \item For existing datasets that are re-packaged, both the original license and the license of the derived asset (if it has changed) should be provided.
        \item If this information is not available online, the authors are encouraged to reach out to the asset's creators.
    \end{itemize}

\item {\bf New assets}
    \item[] Question: Are new assets introduced in the paper well documented and is the documentation provided alongside the assets?
    \item[] Answer: \answerYes{} %
    \item[] Justification: The public repository at \url{https://github.com/xc-j/Event-SAE} provides training configurations, usage instructions, license information, and links to trained SAE checkpoints.
    \item[] Guidelines:
    \begin{itemize}
        \item The answer \answerNA{} means that the paper does not release new assets.
        \item Researchers should communicate the details of the dataset\slash code\slash model as part of their submissions via structured templates. This includes details about training, license, limitations, etc. 
        \item The paper should discuss whether and how consent was obtained from people whose asset is used.
        \item At submission time, remember to anonymize your assets (if applicable). You can either create an anonymized URL or include an anonymized zip file.
    \end{itemize}

\item {\bf Crowdsourcing and research with human subjects}
    \item[] Question: For crowdsourcing experiments and research with human subjects, does the paper include the full text of instructions given to participants and screenshots, if applicable, as well as details about compensation (if any)?
    \item[] Answer: \answerNA{} %
    \item[] Justification: The paper does not involve crowdsourcing or research with human subjects.
    \item[] Guidelines:
    \begin{itemize}
        \item The answer \answerNA{} means that the paper does not involve crowdsourcing nor research with human subjects.
        \item Including this information in the supplemental material is fine, but if the main contribution of the paper involves human subjects, then as much detail as possible should be included in the main paper. 
        \item According to the NeurIPS Code of Ethics, workers involved in data collection, curation, or other labor should be paid at least the minimum wage in the country of the data collector. 
    \end{itemize}

\item {\bf Institutional review board (IRB) approvals or equivalent for research with human subjects}
    \item[] Question: Does the paper describe potential risks incurred by study participants, whether such risks were disclosed to the subjects, and whether Institutional Review Board (IRB) approvals (or an equivalent approval/review based on the requirements of your country or institution) were obtained?
    \item[] Answer: \answerNA{} %
    \item[] Justification: The paper does not involve research with human subjects, so no IRB approval is required.
    \item[] Guidelines:
    \begin{itemize}
        \item The answer \answerNA{} means that the paper does not involve crowdsourcing nor research with human subjects.
        \item Depending on the country in which research is conducted, IRB approval (or equivalent) may be required for any human subjects research. If you obtained IRB approval, you should clearly state this in the paper. 
        \item We recognize that the procedures for this may vary significantly between institutions and locations, and we expect authors to adhere to the NeurIPS Code of Ethics and the guidelines for their institution. 
        \item For initial submissions, do not include any information that would break anonymity (if applicable), such as the institution conducting the review.
    \end{itemize}

\item {\bf Declaration of LLM usage}
    \item[] Question: Does the paper describe the usage of LLMs if it is an important, original, or non-standard component of the core methods in this research? Note that if the LLM is used only for writing, editing, or formatting purposes and does \emph{not} impact the core methodology, scientific rigor, or originality of the research, declaration is not required.
    \item[] Answer: \answerYes{} %
    \item[] Justification: We use a vision-language model (\texttt{gemini-3.1-pro-preview}) to assign phrase and phase labels to event clusters. The labels are visualization aids and do not feed the SAE feature ranking. Details and the prompt template are in Appendix~\ref{app:vlm-prompt}.
    \item[] Guidelines:
    \begin{itemize}
        \item The answer \answerNA{} means that the core method development in this research does not involve LLMs as any important, original, or non-standard components.
        \item Please refer to our LLM policy in the NeurIPS handbook for what should or should not be described.
    \end{itemize}

\end{enumerate}